\documentclass[journal]{IEEEtran}
\usepackage{amsmath,amsfonts}
\usepackage{algorithmic}
\usepackage{algorithm}
\usepackage{array}
\usepackage{textcomp}
\usepackage{stfloats}
\usepackage{url}
\usepackage{verbatim}
\usepackage{graphicx}
\usepackage{textcomp}
\usepackage{cite}

\usepackage{subfigure}
\usepackage{times}
\usepackage{latexsym}

\usepackage{soul}

\usepackage{color}

\usepackage[utf8]{inputenc}

\usepackage{amsmath}
\usepackage{amsthm}
\usepackage{booktabs}

\usepackage{multirow}
\usepackage{xspace}
\usepackage{amsfonts,amssymb}

\usepackage{microtype}

\begin{document}
\title{A Cascade Dual-Decoder Model for Joint Entity \\ and Relation Extraction}
\author{Jian Cheng, Tian Zhang,~Shuang Zhang,~Huimin Ren,~Guo Yu,~\IEEEmembership{Member,~IEEE,}~Xiliang Zhang,~\IEEEmembership{Member,~IEEE} and~Shangce Gao,~\IEEEmembership{Senior Member,~IEEE}, and~ Lianbo Ma,~\IEEEmembership{Senior Member,~IEEE}

\thanks{
\IEEEcompsocthanksitem Jian Cheng is with the Research Institute of Mine Artificial Intelligence in Chinese Institute of Coal Science, State Key Laboratory of Intelligent Coal Mining and Strata Control,  and Tiandi Science and Technology Co., Ltd.,  Beijing 100013, China. E-mail: jiancheng@tsinghua.org.cn
\IEEEcompsocthanksitem Tian Zhang, Shuang Zhang and Huimin Ren are with the College of Software, Northeastern University, Shenyang, China. E-mail: 2190087@stu.neu.edu.cn, 2071348@stu.neu.edu.cn, and renhm01@126.com
\IEEEcompsocthanksitem Yu Guo is with the Institute of Intelligent Manufacturing, Nanjing Tech University, Nanjing, 211816, China. 
E-mail: gysearch@163.com 
\IEEEcompsocthanksitem Xiliang Zhang is with the School of Intelligent Manufacturing and Control Engineering, Shanghai Polytechnic University, Shanghai, China. 
E-mail: xlzhang@sspu.edu.cn
\IEEEcompsocthanksitem Shangce Gao is with the Faculty of Engineering, University of Toyama, Toyama-shi, 930-8555 Japan. Email: gaosc@eng.u-toyama.ac.jp
\IEEEcompsocthanksitem Lianbo Ma is with the College of Software, Northeastern University, Shenyang, China and Foshan Graduate School of Innovation, Northeastern University, Foshan, China. E-mail: malb@swc.neu.edu.cn
\IEEEcompsocthanksitem Corresponding authors: Guo Yu, Lianbo Ma

\IEEEcompsocthanksitem \textcopyright 2024 IEEE.  Personal use of this material is permitted.  Permission from IEEE must be obtained for all other uses, in any current or future media, including reprinting/republishing this material for advertising or promotional purposes, creating new collective works, for resale or redistribution to servers or lists, or reuse of any copyrighted component of this work in other works.
}}


\maketitle

\begin{abstract}
In knowledge graph construction, a challenging issue is how to extract complex (e.g., overlapping) entities and relationships from a small amount of unstructured historical data. The traditional pipeline methods are to divide the extraction into two separate subtasks, which misses the potential interaction between the two subtasks and may lead to error propagation. In this work, we propose an effective cascade dual-decoder method to extract overlapping relational triples, which includes a text-specific relation decoder and a relation-corresponded entity decoder. Our approach is straightforward and it includes a text-specific relation decoder and a relation-corresponded entity decoder. The text-specific relation decoder detects relations from a
sentence at the text level. That is, it does this according to the semantic information of the whole sentence. For each extracted relation, which is with trainable embedding, the relation-corresponded entity decoder detects the corresponding head and tail entities using a span-based tagging scheme. In this way, the overlapping triple problem can be tackled naturally. We conducted experiments on a real-world open-pit mine dataset and two public datasets to verify the method’s generalizability. The experimental results demonstrate the effectiveness and competitiveness of our proposed method and achieve better F1 scores under strict evaluation metrics. Our implementation is available at https://github.com/prastunlp/DualDec.

\end{abstract}


\section{Introduction}

\begin{figure}[t]
  \centering
      \includegraphics[scale=0.65]{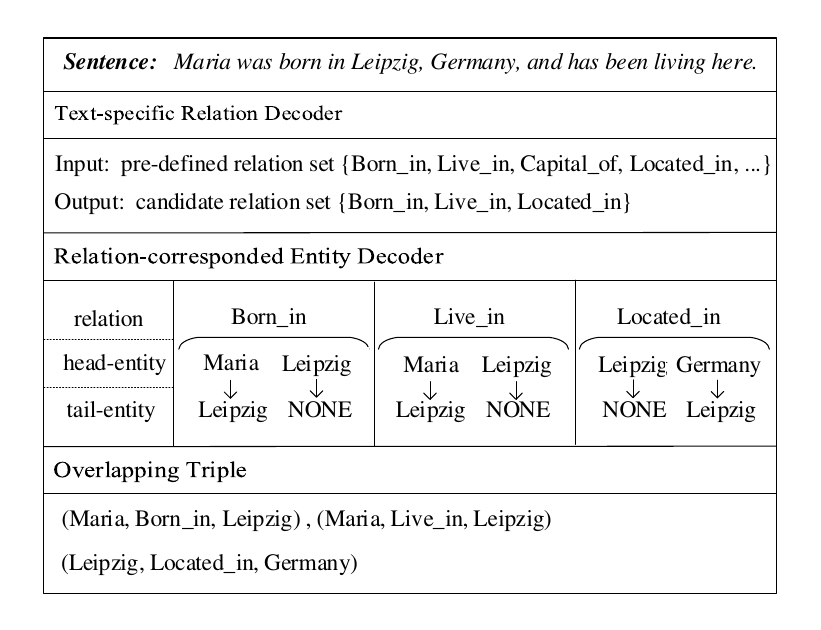}
  \caption{An example to show  the proposed dual-decoder model on the sentence ``{\em Maria was born in Leipzig, Germany, and has been living here}", which contains three overlapping triples.}
  \label{fig:1}
\end{figure}

\IEEEPARstart{E}{xtracting} entities and their semantic relations from unstructured texts is a fundamental task in text mining and knowledge graph construction. Given a text, the aim of this task is to detect relational triples, which consist of two entities and a semantic relation, i.e., in the form of ({\em head entity, relation, tail entity}) or ($h$, $ r$, $t$). The task would become more challenging when a text contains multiple relational triples that share the same entities as shown in Fig. \ref{fig:1}, where the text of ``{\em Maria was born in Leipzig, Germany, and has been living here.}'' contains three relational triples, and both $Born\_{in}$ and $Live\_{in}$ share the same entity pair ({\em Maria, Leipzig}). This is also termed as the overlapping triple problem \cite{DBLP:conf/acl/LiuZZHZ18, DBLP:journals/corr/abs-2011-01675}.

Existing relational triple extraction approaches mainly include pipeline methods and joint extraction methods. Traditional pipeline methods divide the extraction into two separate subtasks, i.e., entity recognition and relation classification. Such methods miss the potential interaction between the two subtasks and may suffer from error propagation since errors in the first stage cannot be remedied in the second stage. Joint extraction methods implement entity extraction and relation extraction in a single model \cite{DBLP:conf/coling/GuptaSA16,DBLP:conf/emnlp/AdelS17,DBLP:conf/acl/KatiyarC17,DBLP:conf/acl/ZhengWBHZX17,DBLP:journals/ijon/ZhengHLBXHX17}, which avoids the error propagation. The unified sequence labeling model \cite{DBLP:conf/acl/ZhengWBHZX17} assigns one tag with entity mention and relation type information to each word for joint entity and relation extraction. But their approach suffers from the problem of overlapping triple due to each token cannot be assigned more than one tag. For better handling the overlapping triple problem, several joint extraction improvements have been developed, such as tagging-based models \cite{DBLP:conf/aaai/DaiXLDSW19,DBLP:conf/coling/WangYZLZS20} and sequence-to-sequence (Seq2Seq)-based models \cite{DBLP:conf/acl/LiuZZHZ18,DBLP:conf/aaai/ZengZL20,DBLP:conf/aaai/NayakN20}. These kinds of methods promote joint model performance improvement and address the problems of pipeline methods initially \cite{ning2023od,gao2023ergm}. However, each step uses an independent decoding algorithm, leading to an accumulation problem of decoding errors, and the redundant errors of the individual modules of the shared parameter integration affect each other's prediction performance, resulting in a cascading redundancy problem \cite{xiong2022multi}. More details can refer to Section \uppercase\expandafter{\romannumeral2}.

Recently, Wei {\em et al}. \cite{DBLP:conf/acl/WeiSWTC20} proposed a new cascade binary tagging framework, called CasRel, which models relations as functions that map head entities to tail entities in a text, rather than discrete labels as in previous works. 
CasRel is a two-stage tagging process: head entities in a text are identified first, and then relations and tail entities are extracted simultaneously. It has been shown that such a cascade tagging framework is a powerful approach, outperforming state-of-the-art methods on different scenarios of overlapping triples. This also motivates us to explore how to utilize potential information of relations as extra features to guide joint extraction since such relations involve somewhat mapping information of entity pairs.

\begin{figure}[t]
  \centering
      \includegraphics[scale=0.65]{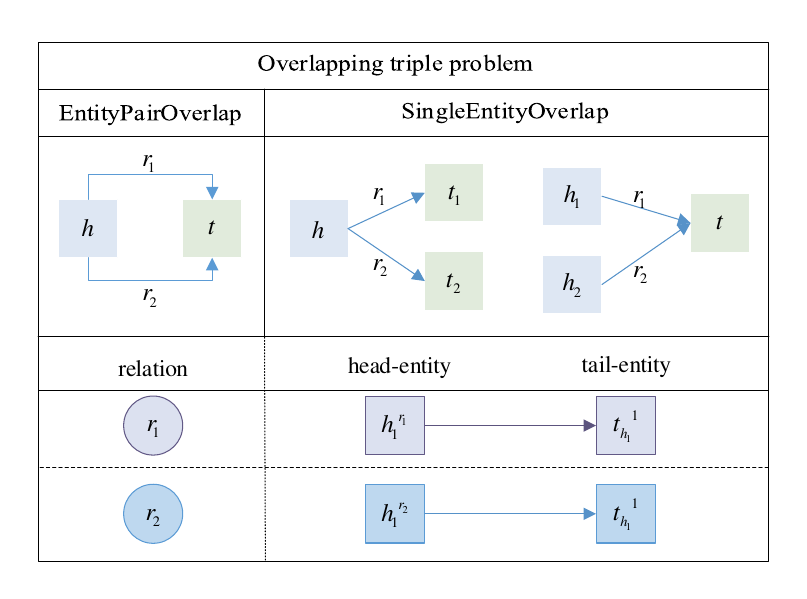}
  \caption{Schematic diagram of the overlapping triple problem. The overlapping triple problem can be tackled  naturally by first detecting the relations and then extracting the entity pairs corresponding to each extracted relation.}
  \label{fig:model_step}
\end{figure}

From the general form of relational triple, we can obtain an intuitive observation: relation is a connection path from head entity to tail entity, reflecting some mapping information between head and tail entities to a certain extent. In other words, once a relation is detected in a text, we can treat it as a mapping function between head and tail entities, rather than a discrete label assigned to an entity pair. Many relation-first extraction models either follow a pipeline approach or a joint extraction method, where relations are typically treated as discrete labels on entity pairs \cite{zhao2021sepc}. The models often treat relations as labels associated with entity pairs, lacking a deeper understanding of the underlying mapping information. The cascade dual-decoder model introduces a novel design \cite{wang2020two,yang2023learning}. Our proposed model has the text-specific relation (TR) decoder and the relation-corresponded entity (RE) decoder. This cascade design allows the model to first detect relations and then extract corresponding entity pairs, providing a more dynamic and interactive approach. And, the model views relations as mapping functions that connect head entities to tail entities, reflecting the inherent mapping information between entities. This perspective guides the cascade extraction process, making it more intuitive and contextually aware. In summary, the proposed cascade dual-decoder model stands out with its innovative approach of viewing relations as mapping functions and leveraging them as extra features for joint extraction. The cascade design and the explicit consideration of overlapping triples contribute to a more context-aware and accurate extraction process.

Accordingly, the entity pairs corresponding to each detected relation can be extracted in a cascade way. In fact, relations are sensitive to contextual semantics since each input text has specific relation information (e.g., number and types of relations). In this sense, it is a feasible way to perform a relation-first cascade extraction approach, that is, we first detect specific relations in a given text, and then treat such text-specific relations as extra knowledge to guide subsequent entity pair extraction process, which can boost the accuracy of entity pair extraction as well as the joint extraction performance. More encouragingly, such a relation-first cascade extraction approach can naturally tackle the overlapping problem (see Fig. \ref{fig:model_step}). For example, in Fig. \ref{fig:1}, a given text ``{\em Maria was born in Leipzig, Germany, and has been living here}'' contains three overlapping triples, i.e., ($Maria, Born\_in, Leipzig$), ($Maria, Live\_in, Leipzig$), and ($Leipzig, Located\_in, Germany$). In this example, a set of candidate relations is first detected from the text, e.g., $\{Born\_in, Live\_in, Located\_in\}$, and then, the potential head and tail entities are identified according to each detected relation. Specifically, for ${Born\_in}$, a head-tail entity pair ``{\em Maria, Leipzig}" can be obtained. For ${Live\_in}$, the extracted entity pair is ``{\em Maria, Leipzig}''. Similarly,  an entity pair ``{\em Germany, Leipzig}'' can be detected according to  ${Located\_in}$. In this way, the overlapping triples can be detected one by one from the text.





\subsection{Summary of Our Approach}
Based on the above motivation, we propose an effective cascade dual-decoder approach for joint entity and relation extraction, as shown in Fig. \ref{fig:2}. It is able to work well in the scenarios of overlapping triples. Our main idea, different from others, is to minimize the difference between the probability distribution of real triples and the extracted ones from our model. By complicated derivation, a formula is acquired, and our model structure is designed straightforwardly according to it. Specifically, in the model, relations are first detected from the text and used as extra mapping features to extract entity pairs; for each detected relation, head entities are predicted and then mapped to the corresponding tail entities. In this way, a common entity or entity pair is allowed to be assigned with different extracted relations, which naturally solves the overlapping triple problem. Moreover, the accuracy of entity recognition in joint extraction can be enhanced by the features of these text relations.

Following the above idea, our model includes a text-specific relation decoder and a relation-corresponded entity decoder. The text-specific relation decoder detects relations from a sentence at the text level, that is,  based on the text semantics of the sentence. For each extracted relation, which is with trainable embedding, the relation-corresponded entity decoder identifies the corresponding head entities and tail entities using a span-based tagging scheme, where start and end positions of entities are tagged. Our method aims to construct an optimized joint decoding algorithm and find an optimal solution to it to obtain the optimal hyperparameters. In this paper, we design a simple and accurate joint extraction framework and enhance the interaction among multiple submodules to attenuate the impact of decoding errors and cascading redundancy on the performance of the joint model due to gradual iteration. We conducted experiments on a real-world open-pit mine dataset and two public datasets to verify the method’s generalizability. The experimental results demonstrate the effectiveness and competitiveness of our proposed method and achieve better F1 scores under strict evaluation metrics. More encouragingly, our proposed approach performs significantly better on relational triple elements.

\subsection{Contributions and Organization}
The main contributions of this paper are as follows:
\begin{itemize}
	\item We propose an effective joint extraction scheme based on the dual-decoder model to address the overlapping triple problem. Our model first detects relations using the text semantics and treats them as extra mapping features to guide entity pair extraction.
	\item We provide the corresponding theoretical analysis to show that our model is mathematically better than recent entity-first cascade models especially on complete entity recognition.
	\item  Extensive experiments are conducted on two public datasets and a real open pit dataset, and the results show that our proposed method and achieve better Fl scores under strict evaluation metrics.
\end{itemize}

The rest of this paper is organized as follows. Section \uppercase\expandafter{\romannumeral2} discusses the related work. The detailed design of the dual-decoder joint extraction model is presented in Section \uppercase\expandafter{\romannumeral3}. Section \uppercase\expandafter{\romannumeral4} conducts comparison experiments on three datasets under different evaluation metrics,  and the conclusion is outlined in Section \uppercase\expandafter{\romannumeral5}.

\section{Related Work}

\subsection{Traditional Joint Models}
Many approaches have been proposed in the literature to extract relational triples from sentences. Early studies in relational triple extraction used pipeline approaches \cite{DBLP:conf/coling/ZengLLZZ14,DBLP:conf/emnlp/GormleyYD15,DBLP:conf/acl/ZhouSTQLHX16,li2023survey} which first recognize all entities in a text and then predict relations of each entity pair. 
A deep convolutional neural network\cite{DBLP:conf/coling/ZengLLZZ14} was adopted to extract lexical and sentence-level features. Yu {\em et al.} \cite{DBLP:conf/ijcai/YuZLWLL19}  proposed a relation extraction model based on segment attention layer  that could learn the  latent relational expressions.
However, these approaches tend to suffer from error propagation and neglect the relevance of entity and relation extraction. In particular, relation classification \cite{DBLP:conf/aaai/FengHZYZ18,DBLP:conf/aaai/ZhangSYLZLG20} can hardly be handled when the same entity participates in multiple relational triples if no sufficient training examples are available. 

To alleviate the pipeline's limitation, latter works proposed to jointly extract entity and relation in a single model, such as feature-based models \cite{ DBLP:conf/acl/LiJ14,DBLP:conf/emnlp/MiwaS14,DBLP:conf/www/RenWHQVJAH17,ma2023pareto}. Generally, these pioneer models were dependent on feature engineering design, which is a labor intensive, manual, time-consuming process. For this issue, benefiting from the fast development of deep learning techniques, recent studies have explored neural network-based models \cite{ DBLP:conf/coling/GuptaSA16,DBLP:conf/acl/KatiyarC17,DBLP:conf/acl/FuLM19,DBLP:conf/aaai/TanZWX19} to reduce manual operations in joint extraction process.
Sun {\em et al.} \cite{DBLP:conf/emnlp/SunWLSWLW18}  proposed a  joint entity and relation extraction model based on the framework of Minimum Risk Training. Bekoulis {\em et al.} \cite{DBLP:journals/corr/abs-1804-07847} transformed the relation extraction task into a multi-head selection problem. Tan {\em et al.} \cite{DBLP:conf/aaai/TanZWX19} performed relation extraction via ranking with translation mechanism.
Note that these traditional neural models \cite{DBLP:conf/acl/MiwaB16} still use the separated decoding and parameters sharing principles to perform joint relation and entity extraction. This means that they still identify relations after recognizing all the entities and thus cannot fully capture the interactions between the two subtasks \cite{DBLP:conf/acl/SunGWGJLSD19}.

\subsection{Tagging-based Models}
Recent tagging-based approaches \cite{DBLP:conf/acl/ZhengWBHZX17,DBLP:conf/aaai/DaiXLDSW19}  have been a popular way. Zheng {\em et al.} \cite{DBLP:conf/acl/ZhengWBHZX17} proposed a unified sequence labeling model that assigns one tag with entity mention and relation type information to each word for joint extraction. But their approach is unable to extract overlapping triples because each token cannot be assigned more than one tag.  For this issue, Takanobu {\em et al.} \cite{DBLP:conf/aaai/TakanobuZLH19} performed a multi-pass tagging on each word based on the reinforcement learning framework.
Then, a novel tagging scheme that runs tagging multiple turns on a text has been introduced \cite{DBLP:conf/aaai/DaiXLDSW19} but this approach is very computationally expensive.

\subsection{Seq2Seq-based Models}
In addition to the above models, some Seq2Seq models \cite{DBLP:conf/acl/LiuZZHZ18,DBLP:conf/emnlp/ZengHZLLZ19, DBLP:conf/aaai/NayakN20} have been utilized for relational triple extraction. Zeng {\em et al}. \cite{DBLP:conf/acl/LiuZZHZ18} proposed a Seq2Seq model using copy mechanism that copies an entity word from sentence directly by decoders. But it is insufficient to extract the full entity mentions. As the improvement,  a sequence-to-sequence model incorporating reinforcement learning that can learn the extraction order of triples and capture the interaction between triples was proposed to generate multiple triples \cite{DBLP:conf/emnlp/ZengHZLLZ19}.
CopyMTL \cite{DBLP:conf/aaai/ZengZL20} was devised to recognize all of the entities in the text by utilizing Conditional Random Field (CRF). As a fixed number (for the triples in a text) is needed for this CopyMTL, its accuracy is compromised when a text contains more than the fixed number of triples.

As stated in \cite{DBLP:conf/acl/WeiSWTC20}, the above existing models all treat relations as discrete labels on entity pairs, which would make joint extraction become a hard machine learning task. In fact, such relations can be modeled as functions that map head entities to tail entities, as shown in the new cascade tagging framework \cite{DBLP:conf/acl/WeiSWTC20}. This also indicates that such relations may involve somewhat mapping information of entity pairs and can be used as extra features to guide joint extraction. Our main focus is not only to model well and exploit this relation as an additional feature for joint extraction, but to provide a new information extraction model for open pit texts using text-specific relation-first cascade extraction principle, which is different from previous works.

\begin{figure*}[t]
  \centering
      \includegraphics[scale=0.8]{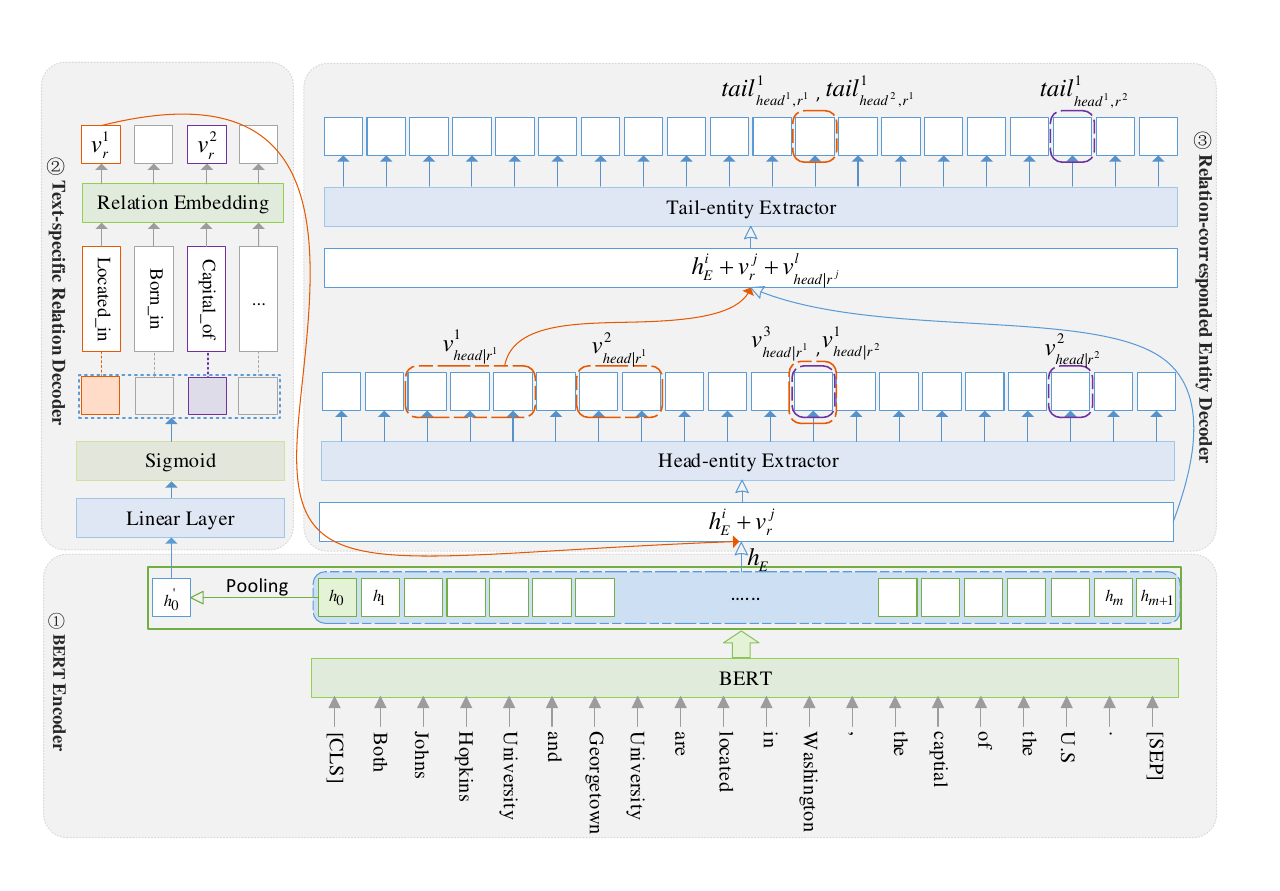}
  \caption{The framework of the cascade dual-decoder model. In this example, there are two candidate relations detected at the TR decoder, i.e., $Located\_in$ and $Capital\_of$. For each candidate relation ${v}_{r}^{j}$, the RE decoder extracts the corresponding head entities and tail entities sequentially. At different relations' iterations ($j$=1, 2), head entities ${v}_{head|r}$ and tail entities ${tail}_{head,r}$ at the RE decoder will change accordingly. A triple is extracted as follows: for instance, for the second relation $j$=2, two candidate head entities, i.e., $Washington$ ($l$=1) and $U.S.$ ($l$=2),  are detected. Then, for $l=1$, a tail entity $U.S.$ is detected, reflecting the triple ($Washington,Capital\_of,U.S$) led by the second candidate relation $Capital\_of$ and the corresponding candidate head entity $Washington$.}
  \label{fig:2}
\end{figure*}

\section{Proposed method}
In this section, we introduce the detailed design of the dual-decoder joint extraction model. As mentioned above, our aim is to minimize the difference between the probability distribution of real triples and the extracted ones from our model. Therefore, our loss function is minimized in Eq. \ref{eq:kl1}:

\begin{equation}
\min_{\theta} {\underset{x}{\mathbb{E}}}\Big\{KL\big(p((h,r,t)|x) \parallel p_{\theta}((h,r,t)|x)\big)\Big\}\label{eq:kl1},
\end{equation}
where $h$, $t$, and $r$ denote head entities, tail entities, and their relations, respectively, $x$ is an annotated text, $\theta$ is the set of trainable parameters, and $KL(p \parallel p_{\theta})$ is the Kullback-Leibler divergence \cite{DBLP:conf/eccv/MathiassenSB02} between the data distribution $p$ and the extracted data distribution $p_{\theta}$ from our model.

\begin{align}
  \nonumber &{\underset{\theta}{min}}{\underset{x}{\mathbb{E}}}\Big\{KL\big(p((h,r,t)|x)||p_\theta((h,r,t)|x)\big)\Big\}\\
 \nonumber  =&{\underset{\theta}{min}}{\underset{x}{\mathbb{E}}}
  \Big\{\int_{(h,r,t)}p((h,r,t)|x) \cdot \\
  \nonumber& \qquad \qquad \log 
  \frac{p((h,r,t)|x)}{p_\theta((h,r,t)|x)}d(h,r,t)\Big\},\\
  \nonumber \Leftrightarrow&{\underset{\theta}{min}}{\underset{x}{\mathbb{E}}}
  \nonumber \Big\{\int_{(h,r,t)}-p((h,r,t)|x)\cdot \\
   &\qquad  \qquad  \log{p_\theta((h,r,t)|x)}d(h,r,t)\Big\}\\
  =&\nonumber \min_{\theta} - {\underset{(h,r,t),x}{\mathbb{E}}}\Big\{\log \big{(}p_{\theta}(r|x)\cdot p_{\theta}(h|r,x) \cdot\\
&\qquad \qquad \qquad \quad p_{\theta}(t|r,h,x)\big{)}\Big\} \label{eq:kl2}.
\end{align}

Through complicated derivation, we acquire Eq. \ref{eq:kl2}, which is the essence of our model structure. More derivation  details are provided in Appendix A. One motivation behind this is that, the text-specific relations can be treated as extra features to guide the joint extraction process. Accordingly, our method first predicts relations using the text semantics to filter out irrelevant ones from the pre-set relation set, and then uses these extra features to boost the recognition of entity pairs. 

From this formula, we can obtain the following observations: 

First, we can treat the three components in Eq. \ref{eq:kl2} as a dual-decoder  procedure, where the first decoder is to learn relations using text semantics, and the second one is to extract entity pairs for each extracted relation. Further, we can regard the second decoder as a cascade tagging procedure (i.e., the last two components in Eq. \ref{eq:kl2}) to extract head and tail entities. Specifically, we first learn a relation classification $p_{\theta}(r|x)$ that identifies relations in a text, and then treat such relations as extra features to recognize entities of the second decoder; for each extracted relation, we learn a head entity extractor $p_{\theta}(h|r,x)$ that predicts head entities from a text with extra information  of the text-specific relation $r$, and then learn a tail entity extractor $p_{\theta}(t|r,h,x)$ to extract the corresponding tail entities. This dual-decoder model is quite different from existing approaches, which usually recognize entities first before joint extraction \footnote{Some recent models \cite{DBLP:conf/acl/LiuZZHZ18,DBLP:conf/aaai/ZengZL20}, such as CopyRe and CopyMTL, also extract relations before identifying entities. However, CopyRe is insufficient to handle full entity mentions and CopyMTL cannot extract more than the fixed number of triples from a text. Obviously, this is not ideal as the actual number of triples is not constant.}. 

Second, compared with the recent entity-first cascade tagging approaches \cite{DBLP:conf/acl/WeiSWTC20}, such a dual-decoder model uses text-specific relations to enhance the accuracy of head/tail entity recognition. This is because that $p_{\theta}(h|r, x)$ (or $p_{\theta}(t|r,h,x)$) is a posterior probability with extra feature information $r$, thus better mathematically than $p_{\theta}(h|x)$ (or $p_{\theta}(t|h,x)$). This point is also supported by the theoretical analysis \footnote{We have given theoretical analysis in Appendix B to illustrate the effectiveness of adding extra information $r$ to predict head entities and tail entities.} and the experiment over relational triple elements (see Table \ref{tab:ele}). 

Third, the second decoder essentially performs the extracted-relation-specific entity pair recognition, which allows a common entity or entity pair to be shared with multiple extracted relations. This is able to solve the overlapping triple problem.

Fig. \ref{fig:2} presents the framework of the cascade dual-decoder model, of which the training objective is to minimize the KL divergence (i.e. Eq.  \ref{eq:kl2}). Given a sentence, the BERT \cite{DBLP:conf/naacl/DevlinCLT19} starts to model its text semantics; the text-specific relation (TR) decoder is to detect potential relations according to the text semantics; and then, for each detected relation, the relation-corresponded entity (RE) decoder extracts the corresponding head and tail entities using a span-based tagging scheme. Details are as follows.


\subsection{BERT Encoder}
We employ the pre-trained BERT model  to encode the text representations (see Fig. \ref{fig:2}) in order to capture the semantics of a text. 

Let $c$ be a special classification token CLS in BERT, and $s$ be a separator SEP in BERT. 
Given a text $x$ with $n$ tokens, the input  to BERT is the concatenation of the tokens as
\begin{equation}
    x = [c,x_1,x_2,...,x_n,s],
\end{equation}
where $x_i$ represents each token in $x$. The BERT will produce an  output $ \mathbf{h}$  with $m+2$ vectors:
\begin{equation}
    \mathbf{h} = [\mathbf{h}_0,\mathbf{h}_1,\mathbf{h}_2,...,\mathbf{h}_m,\mathbf{h}_{m+1}],
\end{equation}
where 
 for $0\leq i\leq m+1$, $\mathbf{h}_i {\in} {\mathbb{R}}^{d}, {d}$ denotes the dimension number of the last hidden layer of BERT,  and  $\mathbf{h}_0$ is the output for $c$ and $\mathbf{h}_{m+1}$ is the output for $s$. For  $1\leq i\leq m$, $\mathbf{h}_i$ is produced by the input of token $x_j$ ($1\leq j\leq n$) \footnote{A token $x_j$ may result in two or
 	  more outputs in $\mathbf{h}_i$, due to stemming, etc. For example, token `packing' will result in two vectors with one for `pack' and another one for `\#\#ing'.}.  

The output of $\mathbf{h}_0$ is used as an input to a pooling layer to produce an input  for the text-specific relation decoder shown in the next section. 
\begin{equation}
	\mathbf{h}_0^{'} = Pooling(\mathbf{h}_0).
\end{equation}

  \textbf {Remarks}: Since BERT is good at capturing context information and word features, which are key to our task, we employ BERT as a base model for the following text-specific relation (TR) decoder and relation-corresponded entity (RE) decoder. To validate our relation-first design, for simplicity, we only use two basic models LinearLayer+sigmoid and LSTM+softmax for TR and RE, respectively. Of course, other complex models can also be applied here to replace the above ones.

\subsection{Text-specific Relation (TR) Decoder}
The number of relations varies in different texts. Therefore, it makes more sense to detect how many relations exist in a text before extracting all the relations and their corresponding triples. As such, detecting the candidate relations in a text is similar to multi-label classification \cite{DBLP:journals/tkde/PengLWWGYLYH21,DBLP:journals/tkde/HuangLHW16,DBLP:journals/tkde/WangPZ20}. 
\subsubsection{Relation detector}
Given a pre-defined set of relations $R = \{r_1,r_2,...,r_K\}$ ($K$ is the size of $R$), the output of $\mathbf{h}_0^{'}$, corresponding to token CLS processed by the pooling layer, is used as input to  identify all the potential relations and it  is fed to  a linear layer before using a sigmoid function $\varphi$ to yield the probability for each relation type:
\begin{equation}
     p_i^r = \varphi(\mathbf{W}^r{\cdot}\mathbf{h}_0^{'} +\mathbf{b}^r),\label{a:r}
\end{equation}
where $\mathbf{W}^r{\in} {\mathbb{R}}^{K \times d}$, $\mathbf{b}^r{\in} {\mathbb{R}}^{K}$.  A text has a relation $r_i$ if $p_i^r$ is higher than a pre-defined confidence threshold $\delta$. For example, as shown in Fig. \ref{fig:2}, the TR decoder detects two candidate relations, i.e., $Located\_in$ and $Capital\_of$.

\subsubsection{Relation embedding} For the relations ($r_{i_1},\dots, r_{i_k}$) detected by the TR decoder from a text $x$, where $k$ is the number of detected relations,  we use the lookup table of relation embeddings to represent each of them (i.e., $r_{i_j}$) as a vector $\mathbf{v}_r^j$ so that such a relation can be used for the next decoder to further extract its head and tail entities. Similarly, the pre-defined set of relations  $R = \{r_1,r_2,...,r_K\}$ can be embedded as $\mathbf{V}_r = [\mathbf{v}_r^1,\dots,\mathbf{v}_r^K]$.

\subsubsection{Optimization  objective of TR}
Given a text representation $\mathbf{x}$, the TR decoder optimizes the following probability to detect the relation $r$:
\begin{align}
p_{\theta}(r|\mathbf{x})=\prod_{i=1}^K (p_{i}^{r})^{y_{i}^{r}} (1-p_{i}^{r})^{1-y_{i}^{r}}\label{eq:r},
\end{align}
where  $y_{i}^{r}$ is the true label of the $i$-th relation.

\begin{figure}[ht]
\centering
    \includegraphics[scale=0.50]{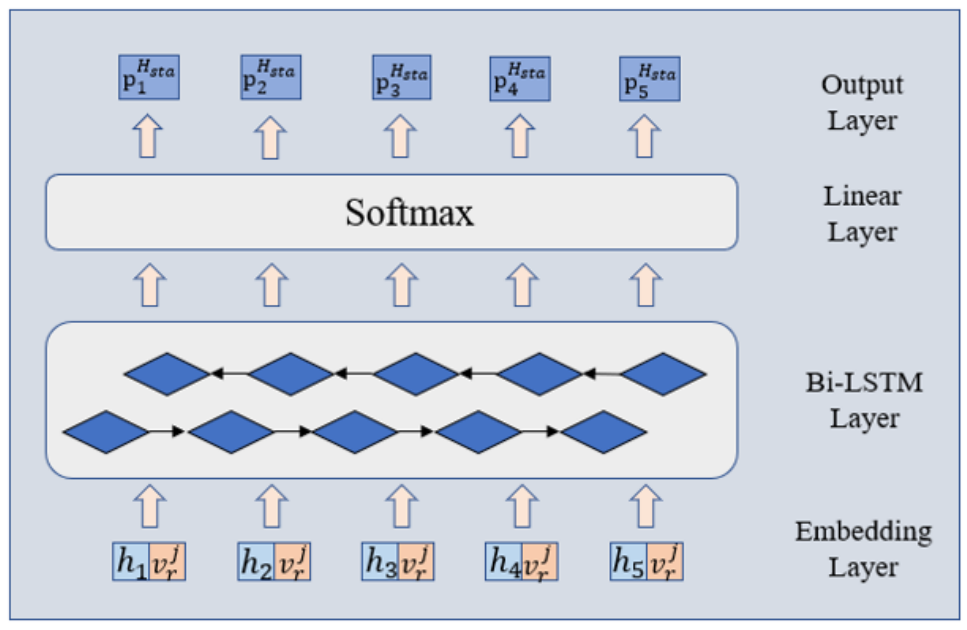}
\caption{Structure of the Head-entity extractor.}
  \label{fig:3}
\end{figure}  

\subsection{Relation-corresponded Entity (RE) Decoder}
The RE decoder includes a head-entity extractor and a tail-entity extractor, as shown in Fig. \ref{fig:2}. 

\subsubsection{Head-entity extractor (HE)} The head-entity extractor aims to detect the corresponding head entities for each extracted relation from the TR decoder. As shown in Fig. \ref{fig:3}, at first, the vectors $\mathbf{h}_i$ and $\mathbf{v}_r^j$ are concatenated and fed into the BiLSTM \cite{DBLP:journals/neco/HochreiterS97} layer to obtain the hidden state of each token's start tag $\mathbf{h}_i^{H_{sta}}$:
\begin{equation}
     \mathbf{h}_i^{H_{sta}} = BiLSTM([\mathbf{h}_i;\mathbf{v}_r^j]),
\end{equation}
where $\mathbf{h}_i$ denotes the final hidden state of each token of BERT and $\mathbf{v}_r^j$ is the $j$-th  candidate relation vector obtained from a lookup table of relation embedding. 

Then, the softmax function   $\sigma(\cdot)$ is used to predict the label of each token as the head entity. Formally, when the start position is tagged, the label of each token is predicted as
\begin{equation}
    p_{i}^{H_{sta}} = \sigma(\mathbf{W}^{H_{sta}} {\cdot} \mathbf{h}_i^{H_{sta}}+\mathbf{b}^{H_{sta}}).\label{a:hs}
\end{equation} 
 
 Analogously,  when the end position is tagged, the label of each token, used as the head entity, can be predicted as
\begin{align}
    &\mathbf{h}_i^{H_{end}} = BiLSTM([\mathbf{h}_i^{H_{sta}};\mathbf{p}_i^{se}])\label{eq:h-h-end}\\
    &p_{i}^{H_{end}} = \sigma(\mathbf{W}^{H_{end}} {\cdot} \mathbf{h}_i^{H_{end}}+\mathbf{b}^{H_{end}})\label{eq:h-end},
\end{align}
where \\
\begin{displaymath}
\mathbf{p}_i^{se}= \left\{ \begin{array}{ll}
i-s^*, & \textrm{if $s^*$ exists}\\
C, & \textrm{otherwise}\\
\end{array} \right.
\end{displaymath}
Here, $s^*$ is the nearest start position in front of the $i$-th token. If $s^*$ exists, $\mathbf{p}_i^{se}$ \cite{DBLP:conf/ecai/0002ZSLWWL20} denotes the vector of relative distance between the $i$-th token and the nearest start position in front of $s^*$. Otherwise, $\mathbf{p}_i^{se}$ is a constant
$C$, normally set to the maximum sentence length \cite{DBLP:conf/ecai/0002ZSLWWL20}.

\subsubsection{Optimization objective of HE}
Accordingly, the head-entity extractor optimizes the following probability to extract the span of head entity $h$:
\begin{align}
p_{\theta}(h|r,\mathbf{x})=\prod_{i=1}^m (p_{i}^{H_{sta}})^{y_{i}^{H_{sta}}} (p_{i}^{H_{end}})^{y_{i}^{H_{end}}}\label{eq:h},
\end{align}
where $m$ is the length of the text, $y_{i}^{H_{sta}}$ and $y_{i}^{H_{end}}$ are the true start and the end tags of head entity for the $i$-th word, respectively.

\subsubsection{Tail-entity extractor (TE)} The tail-entity extractor is to identify the corresponding tail entity for each head entity from the head-entity extractor. When the start position is tagged, it will perform the following operations on each token:
\begin{align}
     &\mathbf{h}_i^{T_{sta}} = BiLSTM([\mathbf{h}_i;\mathbf{v}_r^j;\mathbf{v}_{head}^l])\\
    &p_{i}^{T_{sta}} = \sigma(\mathbf{W}^{T_{sta}} {\cdot} \mathbf{h}_i^{T_{sta}}+\mathbf{b}^{T_{sta}}),\label{a:ts}
\end{align}
where $\mathbf{v}_{head}^l$ represents the vector of the $l$-th head entity from the head-entity extractor, usually defined as the average of the start and end tokens of the $l$-th head entity since a head entity may consist of multiple tokens. 

 When the end position is tagged, the label of each token, used as the tail entity,  is predicted as
\begin{align}
     &\mathbf{h}_i^{T_{end}} = BiLSTM([\mathbf{h}_i^{T_{sta}};\mathbf{p}_i^{se}])\\
    &p_{i}^{T_{end}} = \sigma(\mathbf{W}^{T_{end}} {\cdot} \mathbf{h}_i^{T_{end}}+\mathbf{b}^{T_{end}})\label{eq:t-end} .
\end{align}

\subsubsection{Optimization objective of TE}
Accordingly, the tail-entity extractor optimizes the following probability to extract the span of tail entity $t$:
\begin{align}
p_{\theta}(t|r,h,\mathbf{x})=\prod_{i=1}^m (p_{i}^{T_{sta}})^{y_{i}^{T_{sta}}} (p_{i}^{T_{end}})^{y_{i}^{T_{end}}},\label{eq:t}
\end{align}
where $y_{i}^{T_{sta}}$ and $y_{i}^{T_{end}}$ are the true start and end tags of tail entity for the $i$-th word, respectively.

For the tail-entity extractor, the result of detecting tail entities is decided by the outputs of relation-decoder and head-entity extractor in a cascade way. For example, as shown in Fig. \ref{fig:2}, the tail entity ``{\em Washington}" is led by the detected relation ``$Located\_in$" and the candidate head entity ``{\em Johns Hopkins University}" (or ``{\em Georgetown University}"). Such a cascade way can detect different head-tail entity pairs under specific relations from relation-decoder, which can naturally deal with the overlapping problem.

\subsection{Joint Training}
To train the model in a joint manner, we rewrite our loss function of Eq. \ref{eq:kl2} into Eq. \ref{eq:loss}:
\begin{equation}
\begin{aligned}
    \mathcal{L}  =& -\underset{(h,r,t),x}{\mathbb{E}} \big \{\log p_{\theta}(r|\mathbf{x})+ \\
    & \log  p_{\theta}(h|r,\mathbf{x})
    + \log  p_{\theta}(t|r,h,\mathbf{x}) \big \} \label{eq:loss},
\end{aligned}
\end{equation}
where $p_{\theta}(r|\mathbf{x})$ , $p_{\theta}(h|r,\mathbf{x})$  and  $p_{\theta}(t|r,h,\mathbf{x})$ are defined in Eq. \ref{eq:r}, Eq. \ref{eq:h} and
Eq. \ref{eq:t}, respectively.

\begin{algorithm}[tb]
\caption{Inference Algorithm}
\label{alg:algorithm}
\textbf{Input}: a text $X$\\
\textbf{Output}: a relational triple set {\bf{R}}
\begin{algorithmic}[1] 
\STATE Initialize {\bf{R}} ${\leftarrow}$ $\{\}$ \\
\STATE Defined $n$ ${\leftarrow}$ Text Length\\
\STATE Obtain $relation\_set(r)$ by Eq. \ref{a:r}\\
\STATE Defined $m$ ${\leftarrow}$ the size of $relation\_set(r)$ \\
\FOR{$idr$ $ {\leftarrow}$  1 to $m$ } \do\\
\STATE $r$ $=$ $relation\_set(r)[idr]$ \\
\STATE Compute $h\_sta(h)$ by Eq. \ref{a:hs}\\
\STATE Compute $h\_end(h)$ by Eq. \ref{eq:h-end}\\
\FOR{$idh_s$ ${\leftarrow}$ 1  to $n$} \do \\
\IF{ $h\_sta(h)[idh_s]$ != 'O'} 
\FOR{ $idh_e$ ${\leftarrow}$ $idh_s$  to n }  \do \\
\IF{$h\_end(h)[idh_e]$ = $h\_sta(h)[idh_s]$} 
\STATE $h = X[idh_s : idh_e]$ \\
\STATE Compute $t\_sta(t)$ by Eq. \ref{a:ts} \\
\STATE Compute $t\_end(t)$ by Eq. \ref{eq:t-end} \\
\STATE Obtain $idt_s$ and $idt_e$ using the same steps as Lines 9-12 \\
\STATE $t = X[idt_s: idt_e]$ \\
\STATE $\bf{R}$ ${\leftarrow}$ $\bf{R}$ $\cup$  $\{(h, r, t)\}$ \\
\STATE Break\\
\ENDIF
\ENDFOR
\ENDIF
\ENDFOR
\ENDFOR
\STATE \textbf{return} {\bf{R}}\\
\end{algorithmic}
\end{algorithm}

\subsection{Inference Algorithm}

\begin{figure}[ht]
  \flushleft 
    \includegraphics[scale=0.55]{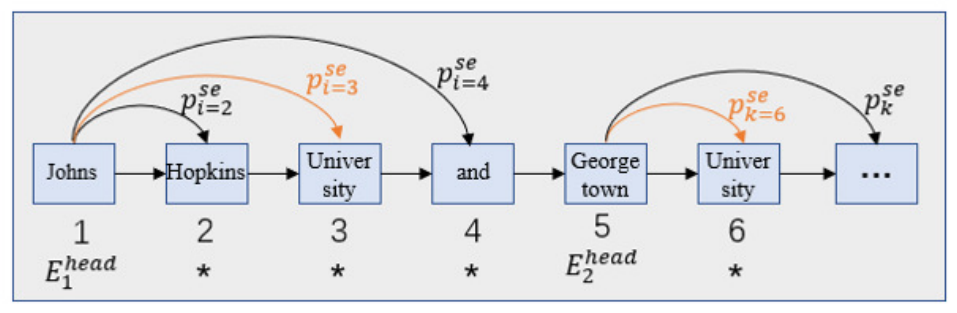}
\caption{The matching process of an entity's start position and end position. $\mathbf{E}_i^{head}$ denotes the start position of the ith entity, i$\in$[1,c], c denotes the number of entities in a sentence. * denotes the possible end position of the i-th entity. $\mathbf{p}_i^{se}$ denotes the relative distance between the start position and the end position of the entity.}
  \label{fig:4}
\end{figure}

At the inference time, we propose a straightforward decoding algorithm for adapting to complex joint extraction tasks, as shown in Algorithm \ref{alg:algorithm}.
For each input text $X$, we initialize {\bf{R}} as an empty set to record extracted triples (Line 1).
Next,  $n$ is defined as the length of the text $X$ (Line 2),  the candidate relation set ${relation\_set(r)}$ is obtained by Eq.\ref{a:r} (Line 3) and $m$ is defined as the size of $relation\_set(r)$ (Line 4). 
Then, we can obtain the tag sequences $h\_sta(h)$ (i.e., start positions) and $h\_end(h)$ (i.e., end positions) of head entities using Eq. \ref{a:hs} and Eq. \ref{eq:h-end}, respectively (Lines 7-8). Similarly, the tag sequences ($t\_sta(t)$ and $t\_end(t)$) of tail entities can be computed using Eq. \ref{a:ts} and Eq. \ref{eq:t-end}, respectively (Lines 14-15). The matching process of the start position and end position for an entity is shown in Fig. \ref{fig:4}. Based on the above computation, the detailed decoding process is as follows: 

First, we traverse $relation\_set(r)$ to get a candidate realtion $r$ (Line 6), and then traverse $h\_sta(h)$ to find the start position of a head entity $h$: if $idh_s$ (current index) is not “O” (non-entity), which indicates that this position may be a start token (Line 10), we need to traverse $h\_end(h)$ from $idh_s$ to search for an end position  $idh_e$ (Line 11). In this procedure, we use a criterion to match the start position with the end position. That is, if the tag of the index $idh_e$  is identical to the index $idh_s$ (Line 12), the tokens between the two indices are considered to be a head entity $h$ (Line 13). Next, we can also traverse $t\_sta(t)$ and $t\_end(t)$ to find a tail entity $t$ (Line 16) , which is similar to the process of searching for head entity $h$. 
Then, the extracted triple $(h,r,t)$ is added into {\bf{R}}.  
\subsection{Discussion}
Our framework is derived from a new formula  Eq. \ref{eq:kl2}, essentially different from other relation-first models like CopyRe \cite{DBLP:conf/acl/LiuZZHZ18}. First, authors of CopyRe only give a loss function without any mathematical explanation. But we infer that they tried to maximize the mean likelihood of each word, which might not be a correct way, since it might suffer from incorrect prediction of entities with more than one word. Furthermore, it might not distinguish between head entities and tail ones correctly. While ours can avoid these problems since ours considers the probability distribution of all triples and minimizes the difference between real triples and extracted ones, which is essentially different and more reliable (cf. Appendix). Finally, ours treats text-specific relations as implicit functions that map head entities to tail entities, different from (Seq2Seq)-based CopyRe.

\begin{table}[t]
\caption{ Statistics of the used datasets. $( )$ refers to the number of the sentences where the complete entity names are annotated; others refer to the cases where only the last words of the entity names are annotated.}
\label{tab:dataset}
    \small
\centering
\begin{tabular}{lcccccc}
\toprule
\multicolumn{1}{c}{\multirow{2}*{Category}}  & \multicolumn{2}{c}{NYT} & \multicolumn{2}{c}{WebNLG} & \multicolumn{2}{c}{MDD} \\
\cmidrule(lr){2-3} \cmidrule(lr){4-5}  \cmidrule(lr){6-7} 
                &   Train & Test    & Train & Test & Train & Test \\
\midrule 
  
\multirow{2}*{ \#Normal}    & 37013  & 3266 & 1596 & 246 &  &  \\
                                & (35703)  & (3136) & (1859) & (254) & (517) & (78) \\
\midrule
\multirow{2}*{ \#EPO}       & 9782   & 978 & 227 & 26 &  & \\
                                & (11991)   & (1168) & (40) & (6) & (113) & (16) \\
\midrule
\multirow{2}*{ \#SEO}       & 14735  & 1297 & 3406 & 457 &  & \\
                            & (14374)  & (1273) & (3154) & (448) & (687) & (101) \\
\midrule
\multirow{1}*{ \#ALL}       & 56195 &  5000 & 5019 &  703  & 1145 &  165\\

\multirow{1}*{ \#Relation}  & \multicolumn{2}{c}{24} & \multicolumn{2}{c}{246} & \multicolumn{2}{c}{12} \\
\bottomrule
\end{tabular}
\end{table}

\begin{table*}[t]
\caption{Results on triple extraction under {\em Partial Match} and {\em Exact Match}. Results marked by $\dagger$ are reported from ~\protect\cite{DBLP:conf/ijcai/LiuCWZLX20}, results marked by $*$ are reproduced by us, and other results are reported by previous models.} 
\label{tab:res}
    \small
\centering
\begin{tabular}{lccccccccccccccc}
\toprule
\multirow{3}*{Methods}   &\multicolumn{6}{c}{\em{Partial Match}}    & \multicolumn{9}{c}{\em{Exact Match}}\\
\cmidrule(lr){2-7} \cmidrule(lr){8-16}
    &\multicolumn{3}{c}{NYT(N)}  &\multicolumn{3}{c}{WebNLG(W)}         &\multicolumn{3}{c}{NYT(N)} &\multicolumn{3}{c}{WebNLG(W)} &\multicolumn{3}{c}{MDD}\\
\cmidrule(lr){2-4} \cmidrule(lr){5-7} \cmidrule(lr){8-10} \cmidrule(lr){11-13} \cmidrule(lr){14-16}
                                     &Pre.&Rec.&F1    &Pre.&Rec.&F1                 &Pre.&Rec.&F1   &Pre.&Rec.&F1     &Pre.&Rec.&F1 \\
\midrule
NovelTagging  & 62.4 & 31.7 &  42.0 &  52.5 &  19.3 & 28.3    &  - &  - & - &  - &  - & - &  49.64 &  50.88 & 50.25\\
CopyRe   & 61.0 & 56.6 &  58.7 &  37.7 &  36.4 & 37.1    &  - &  - & - &  - &  - & -  &49.37	&52.64	&50.95 \\
GraphRel             & 63.9 & 60.0 &  61.9 &  44.7 &  41.1 & 42.9    &  - &  - & - &  - &  - & -  &52.03	&51.89	&51.96 \\
CopyRL         & 77.9 & 67.2 &  72.1 &  63.3 &  59.9 & 61.6    &  - &  - & - &  - &  - & -  &71.53	&69.73	&70.62\\
CopyMTL             &  - &  - & - &  - &  - & -                     &  75.7 &  68.7 & 72.0 &  58.0 &  54.9 & 56.4  &51.36	&48.90	&55.10\\
WDec                  &  - &  - & - &  - &  - & -                     &  88.1 &  76.1 & 81.7 &  88.6$^\dagger$ &  51.3$^\dagger$ & 65.0$^\dagger$  &66.74	&50.06	&57.21\\
AttentionRE      &  - &  - & - &  - &  - & -                     &  88.1 &  78.5 & 83.0 &  \bf{89.5} &  86.0 & 87.7 &68.94	&69.20	&69.07\\
CasRel      & 89.7 & 89.5 &  89.6 &  \bf{93.4} &  90.1 & \bf{91.8}    &  89.1$^*$ & 89.4$^*$ & 89.2$^*$ &  87.7$^*$ &  85.0$^*$ & 86.3$^*$  &71.53	&69.73	&70.62\\
SAHT & 83.4 & 80.6 & 81.9 & 85.8 &  81.9 & 83.9 &- &- &\- &- &-&- &65.01	&61.91	&63.42\\

MAGCN & 85.2 & 59.7 &\ 70.2 &84.6 & 50.3 &\ 63.1 &- &- &- &- &-&- &62.24	&58.15	&60.13\\
Ours  & \bf{90.2} & \bf{90.9} &  \bf{90.5} &  90.3 &  \bf{91.5} & 90.9    &  \bf{89.9} &  \bf{91.4} & \bf{90.6} &  88.0 &  \bf{88.9} & \bf{88.4}  &\bf{73.48}	&\bf{71.28}	&\bf{72.36}\\
\bottomrule
\end{tabular}

\end{table*}




\section{Experiment}
\subsection{Datasets}
Two public datasets are used in our evaluation: NYT \cite{DBLP:conf/pkdd/RiedelYM10} and WebNLG \cite{DBLP:conf/acl/GardentSNP17}. NYT, derived from the distant supervised relation extraction task, contains 24 relation types with 5000 sentences for test, 5000 sentences for validation set, and 56195 sentences for training. WebNLG, originally used for the natural language generation task, includes 246 relation types with a training set of 5019 sentences, a test set of 703 sentences, and a validation set of 500 sentences. Both of these datasets involve the overlapping relational triples. According to the overlapping triple patterns \cite{DBLP:conf/acl/LiuZZHZ18}, we classify the test set into three types: {\em Normal}, where the triples of a sentence do not share any entities; {\em EntityPairOverlap} ({\em EPO}), where at least two triples in a sentence are overlapped by one entity pair; {\em SingleEntityOverlap} ({\em SEO}), where a sentence has at least two triples that share exactly one entity. The statistics of the above datasets are shown in Table \ref{tab:dataset}.

In addition, we construct a real open mine dataset from scratch, i.e., Mine Disaster Dataset (MDD). The source of the original corpus is built on publicly available mine data sites, and the data is selected and filtered by hand. Each piece of data is annotated by well-educated annotators, including the labling of entities and relations between entities, where the incorrectly labeled sentences are filtered out. The MDD dataset contains 1638 samples with 12 mine entities and 12 relationship categories, as shown in Table \ref{tab:dataset2}. We divide the dataset into training set, validation set and test set using 1145, 328 and 165 sentences, respectively.

\begin{table}[t]

    \small
\centering
\caption{ Predefined categories in the dataset.}
\label{tab:dataset2}
\begin{tabular}{ll} 
\toprule
\multicolumn{1}{c}{\multirow{1}*{}} & \multicolumn{1}{c}{Categories} \\
\midrule 
\multirow{1}*{Entity type}    &name of mine, geology, geological characteristics, \\
\multirow{1}*{}    &features, landslide causes, management measures, \\
\multirow{1}*{}    &mine location, detailed management measures, \\
\multirow{1}*{}    & restrictions, hydrological conditions \\
\multirow{1}*{}    &  accident types,accident specific forms. \\
\midrule
\multirow{1}*{Relation}  &accident type, inclusion, characteristics,     \\
\multirow{1}*{}    &accident causes, preventive measures, lithology,\\
\multirow{1}*{}    &specific location, maximum restrictions,   \\
\multirow{1}*{}    & rock stratum, accident specific type,   \\
\multirow{1}*{}    & treatment measures, specific methods.  \\
\bottomrule
\end{tabular}

\end{table}

\subsection{Evaluation Metrics}
We employ two evaluation metrics for comprehensive experiment studies: (1) {\em Partial Match} \cite{DBLP:conf/acl/LiuZZHZ18}, where an extracted triple ($h$, $r$, $t$) is regarded as correct only if its relation and the last word of the head entity name and the tail entity name are correct. Note that only the last word of an entity name in both training and test sets is annotated; (2) {\em Exact Match} \cite{DBLP:conf/aaai/NayakN20,DBLP:conf/ijcai/LiuCWZLX20}, where a predicted triple is regarded as correct only if its relation and the full names of its head and tail entities are all correct. In this metric, the complete entity names are annotated. For a fair comparison, three standard evaluation indicators, i.e., precision (Pre.), recall (Rec.) and F1 score (F1), are used to evaluate the experimental results.

\subsection{Implementation Details}

Following CasRel  \cite{DBLP:conf/acl/WeiSWTC20}, we employ the PyTorch version of BERT$_{base}$(cased) \footnote{Available at: https://github.com/huggingface/transformers.} as the pre-trained BERT model. For the text-specific relation decoder, we use relation embeddings in GloVe word embedding glove.840B.300d \cite{DBLP:conf/emnlp/PenningtonSM14} with 300 dimensions  \cite{DBLP:conf/acl/FuLM19}. For the relation-corresponded entity decoder, the hidden dimensionality of the forward and the backward LSTM  is set to be 384 \cite{DBLP:conf/acl/ZhengWBHZX17,DBLP:conf/aaai/NayakN20,DBLP:conf/acl/FuLM19, zhao2022subject,tao2023multi}. The threshold for the text-specific relation decoder is set to be 0.5. Following \cite{DBLP:conf/acl/LiuZZHZ18,DBLP:conf/emnlp/ZengHZLLZ19,DBLP:conf/aaai/ZengZL20,DBLP:conf/aaai/NayakN20}, we use the Adam optimizer during the model training. The learning rate and the batch size are set to be 2e-5 and 8, respectively. All the hyper-parameters are tuned on the validation set. To avoid overfitting, we apply dropout at a rate of 0.4. All experiments are conducted with an NVIDIA GeForce RTX 2080 Ti. Furthermore, the experiments take 10 hours on the NYT dataset for 70 epochs, 14 hours on the WebNLG dataset for 70 epochs. And, the experiments take 10 seconds to do the inference task on the dataset with 5000 entries.

\subsection{Baseline Methods}
We compare our method on NYT and WebNLG datasets with the following baselines: 
\begin{itemize}
\item {\bf NovelTagging} \cite{DBLP:conf/acl/ZhengWBHZX17} is the first unified sequence labeling model, which predicts both entity type and relation class for each word.
\item {\bf CopyRE} \cite{DBLP:conf/acl/LiuZZHZ18} is a neural encoder-decoder structure for extracting the overlapping triples. We report the results of the MultiDecoder.
\item {\bf GraphRel} \cite{DBLP:conf/acl/FuLM19} is a two-phase model based on graph convolution network (GCN), which considers each token as a node in a graph and regards the edge connecting to two nodes as a relation between them.
\item {\bf CopyRL} \cite{DBLP:conf/emnlp/ZengHZLLZ19} applies the reinforcement learning into a sequence-to-sequence model to generate multiple triplets, which can learn the extraction order of triplets and capture interactions among them.
\item {\bf WDec} \cite{DBLP:conf/aaai/NayakN20} introduces a novel approaches using encoder-decoder architecture to generate triplets with entire boundaries.  
\item  {\bf CopyMTL} \cite{DBLP:conf/aaai/ZengZL20} uses a multi-task learning framework equipped with copy mechanism, which allows the model to extract multi-token entities.
\item  {\bf AttentionRE} \cite{DBLP:conf/ijcai/LiuCWZLX20} is a joint extraction model with the supervised multi-head self-attention mechanism, which contains an entity extraction module and a relation detection module.
\item  {\bf CasRel} \cite{DBLP:conf/acl/WeiSWTC20} is a state-of-the-art method using novel cascade binary tagging framework to extract multiple relational triples, which has achieved promising performance on NYT and WebNLG datasets.

\item  {\bf SAHT} \cite{zhao2022subject} is a subject-aware attention-based hierarchical tagger used to extract both entities and relationships from unstructured text. 
\item  {\bf MAGCN} \cite{tao2023multi} is an entity relationship joint extraction model using a multi-head attention graph convolutional network. i.e., a network model is constructed using a multi-head attention mechanism to generate an adjacency matrix and identify multiple relationships by head selection.

\end{itemize}

\subsection{Main Results}

For a fair comparison, we employ the published results of previous studies. Table \ref{tab:res} shows the performance of our proposed method against the baselines on NYT and WebNLG datasets in terms of {\em Partial Match} and {\em Exact Match}, where most results have been reported previously. From this table, we observe that our model and CasRel achieve significantly better results than other baselines on the two datasets. For {\em Partial Match}, our model obtains the best F1 score (90.5\%) (with at least 18.4\% improvement over NovelTagging, CopyRe, GraphRel, MAGCN, and CopyRL. Moreover, our method outperforms SAHT by 6.8\%,10.3\%, 8.6\% in precision, recall, and F1 score.) on NYT, and the second best F1 score (90.9\%) on WebNLG. For {\em Exact Match}, which is a stricter metric, our model shows an obvious performance advantage over the baselines on the two datasets. Our model achieves the best precision (89.9\%), recall (91.4\%) and F1 score (90.6\%) (with 1.4\% and at least 7.6\% improvements over CasRel and other baselines, respectively) on NYT. On WebNLG, our model performs better than all the baselines, achieving the best F1 score. Besides, we also compare our model with other baselines (NovelTagging, CopyRe, GraphRel, CopyRL, CopyMTL, WDec, AttentionRE, SAHT, MAGCN and CasRel) on MDD dataset. As shown in Table \ref{tab:res}, the NovelTagging, CopyRe, GraphRel, CopyRL, CopyMTL, WDec, and AttentionRE methods perform poorly on the MDD dataset compared with CasRel and our model. Our model outperforms CasRel in F1 score by 1.74\% and performs better than CasRel in terms of precision (1.95\% higher) and recall (1.55\% higher). These encouraging results show that our model is more powerful in dealing with complex extraction tasks where the complete name rather than the last word of the entity needs to be identified. This also validates the effectiveness of the proposed cascade dual-decoder (i.e., TR and RE decoders) model. 

\subsection{Ablation studies}
\subsubsection{Result Analysis on Different Triple Elements} We further explore the performance of our model on relational triple elements. Table \ref{tab:ele} shows the comparison between our model and the best baseline CasRel on NYT and WebNLG datasets, where an element (e.g., $(h, t)$) is regarded as correct only if its involved items (e.g., $h$ and $t$) are all correct, regardless of the correctness of the other items (e.g., $r$) in the extracted triple $ (h, r, t)$. From Table \ref{tab:ele}, it is observed that our model outperforms the baseline on most test instances, achieving better F1 scores on relational triple elements in terms of {\em Partial Match} and {\em Exact Match}. 

\begin{table}[t]
\caption{ F1 score results on relational triple elements in terms of {\em Partial Match} and {\em Exact Match}.}
\label{tab:ele}
	\small
\centering
\begin{tabular}{lccccc}
\toprule
\multirow{2}*{Methods}  & \multirow{2}*{Element}  & \multicolumn{2}{c}{\em{Partial Match}} & \multicolumn{2}{c}{\em{Exact Match}}\\
\cmidrule(lr){3-4} \cmidrule(lr){5-6} 
                &   & N & W    & N & W  \\
\midrule 
  
\multirow{3}*{CasRel}   &  $ (h, t)$      &  89.7 &  \bf{93.5} & 89.7 &  87.6  \\
                        &  $ r$          &  94.9 &  94.0 & 95.2 &  93.7  \\
                        &  $ (h, r, t)$    &  89.6 &  \bf{91.8} & 89.2 &  86.3 \\
\midrule

\multirow{3}*{Ours}     &  $ (h, t)$      &  \bf{90.7} &  \bf{93.5} & \bf{90.8} &  \bf{90.9}  \\
                        &  $ r$            &  \bf{95.6} &  \bf{94.4} & \bf{95.6} &  \bf{94.6}  \\
                        &  $ (h, r, t)$   &  \bf{90.5} &  90.9 & \bf{90.6} &  \bf{88.4} \\
\bottomrule
\end{tabular}
\end{table}

We also find that there is a gap between the performance on $r$ and $ (h, t)$ on NYT, WebNLG and MDD for the two methods, which reveals that extracting entity pairs is more challenging than identifying relations from a text. In a sense, we can use the relations that can be easily identified from a text as extra features to guide the following complex entity pair extraction. This also supports one motivation of our relation-first design that highlights the effect of text-specific relation information on entity recognition, while CasRel tags the start and end words of tail entities using the pre-defined relation type, which does not make sufficient use of relation information. Note that Table \ref{tab:ele} further shows that our model achieves an obvious performance improvement over CasRel under the strict metric {\em Exact Match}. This indicates that our model is good at recognizing entity pair $(h, t)$ in the complete entity extraction scenario, which also reveals the model's real value in practical applications.

Notably, for WebNLG, the performance gap between $ (h,t) $ and $ (h, r, t)$ is comparatively larger than NYT. It indicates that it is harder to identify relations in WebNLG. We think that it is because the number of relations contained in WebNLG is 246 while NYT contains 24 relations so the relation identification is more difficult. As shown in Table \ref{tab:ele}, 
our model performs more powerfully than CasRel on WebNLG, especially under strict metric {\em Exact Match}. The reasons may be as follows: compared to CasRel, our model works in a relation-first way, which can take a better utilization of the text-specific relation information in the joint extraction process. Especially, our model first detects all the potential relations for a specific text (i.e., this is based on the text level) and then uses such relation information to guide to extraction of the corresponding entity pairs of each relation. In this sense, for entity extraction, CasRel only uses the information about the relation type to tag the start and end words of tail entities, while ours explicitly incorporates the relation embedding, which can provide more useful information to guide the entity extraction process.

\subsubsection{Result Analysis on Different Sentence Types} 
\begin{figure*}[htb]
    \centering
        \subfigure[Normal]{
        \begin{minipage}[t]{0.3\linewidth}
        \centering
        \includegraphics[scale=0.3]{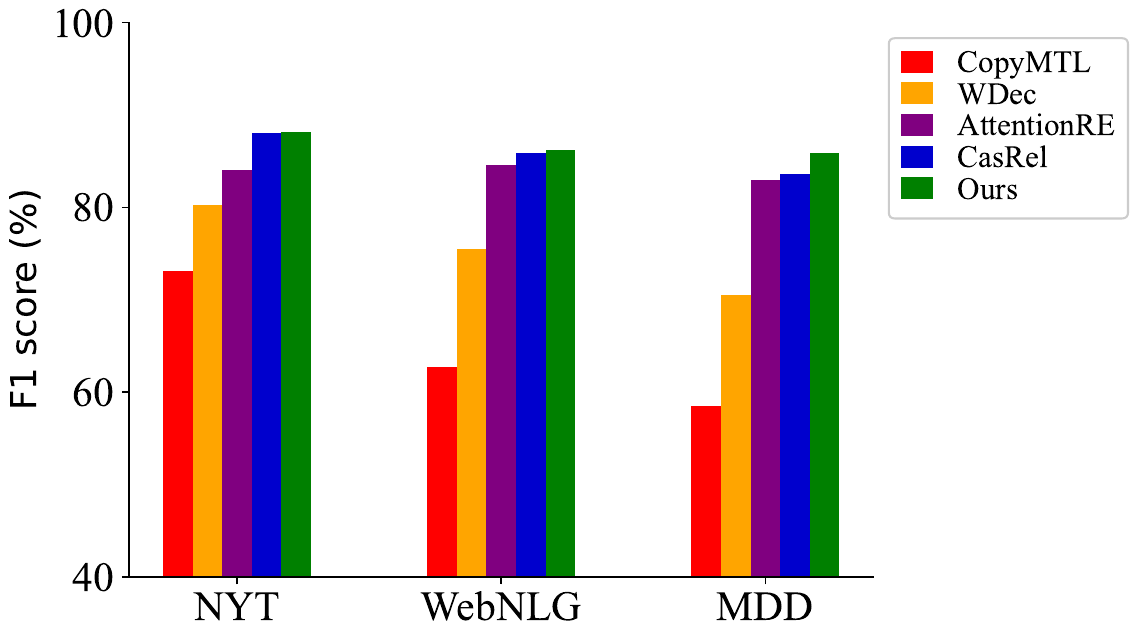}
        \end{minipage}%
        }%
        \subfigure[EntityPairOverlap ]{
        \begin{minipage}[t]{0.3\linewidth}
        \centering
        \includegraphics[scale=0.3]{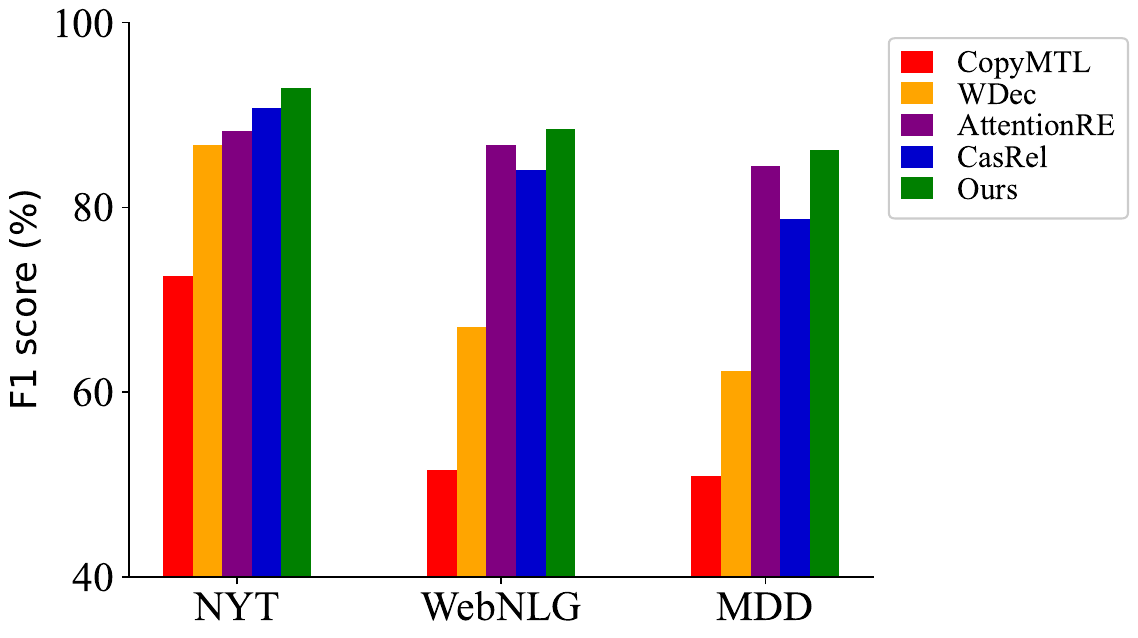}
        \end{minipage}%
        }%
        \subfigure[SingleEntityOverlap]{
        \begin{minipage}[t]{0.3\linewidth}
        \centering
        \includegraphics[scale=0.3]{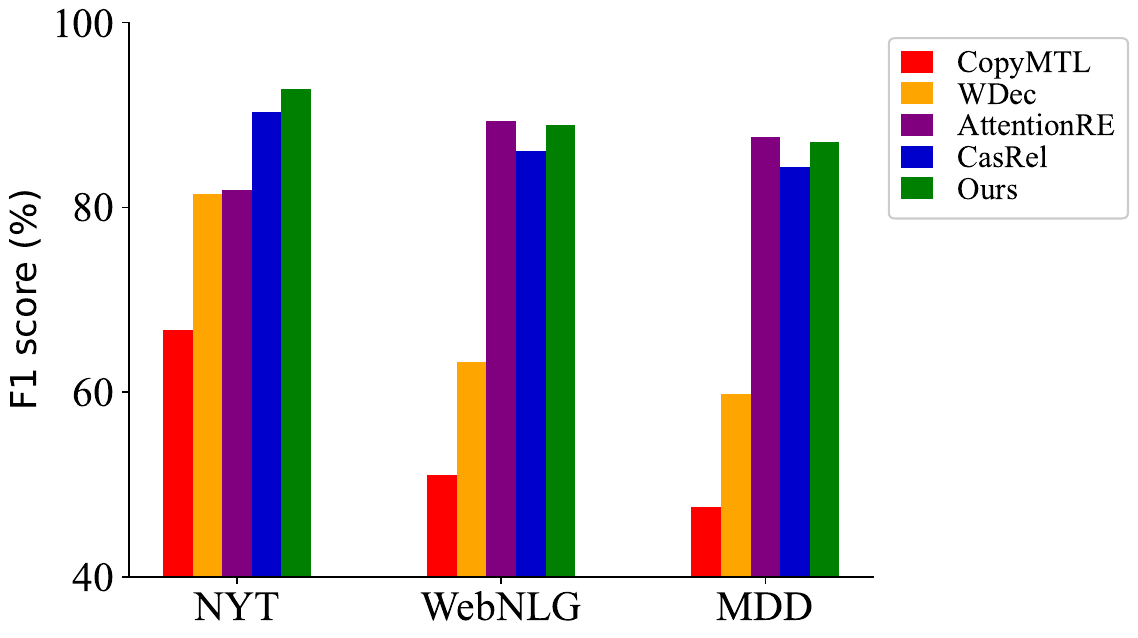}
        \end{minipage}%
        }
    \caption{Results on different sentence types according to the overlapping degree in terms of {\em Exact Match}.}
    \label{fig:type}
\end{figure*}

To investigate the performance of our model in extracting overlapping relational triples, we further analyze the statistics obtained by our proposed model and four baselines (i.e., CasRel, CopyMTL, WDec and AttentionRE) on different types of sentences in terms of {\em Exact Match}. Fig. \ref{fig:type} shows the detailed results on the three overlapping patterns, where {\em Normal} is the easiest pattern while {\em EPO} and {\em SEO } are more difficult to be handled. It can be observed that our model gains consistently best performance over three overlapping patterns. Especially for those hard patterns {\em EPO} and {\em SEO}, our model achieves better F1 scores than the best baseline CasRel, which shows its promising capability in dealing with the overlapping triple problem.

\begin{figure*}[htb]
    \centering
        \subfigure[NYT]{
        \begin{minipage}[t]{0.3\linewidth}
        \centering
        \includegraphics[scale=0.3]{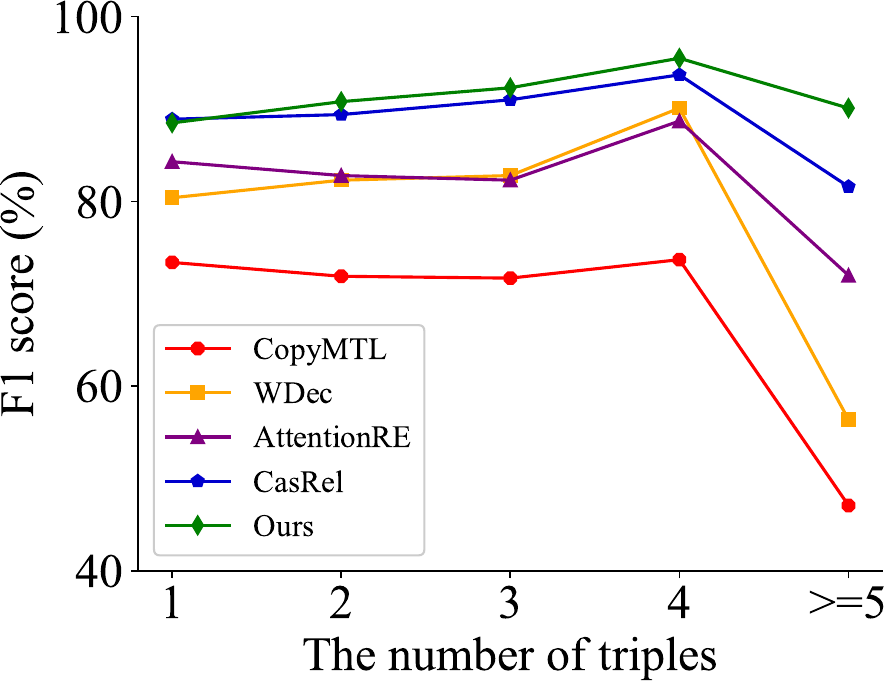}
        \end{minipage}%
        }%
        \subfigure[WebNLG]{
        \begin{minipage}[t]{0.3\linewidth}
        \centering
        \includegraphics[scale=0.3]{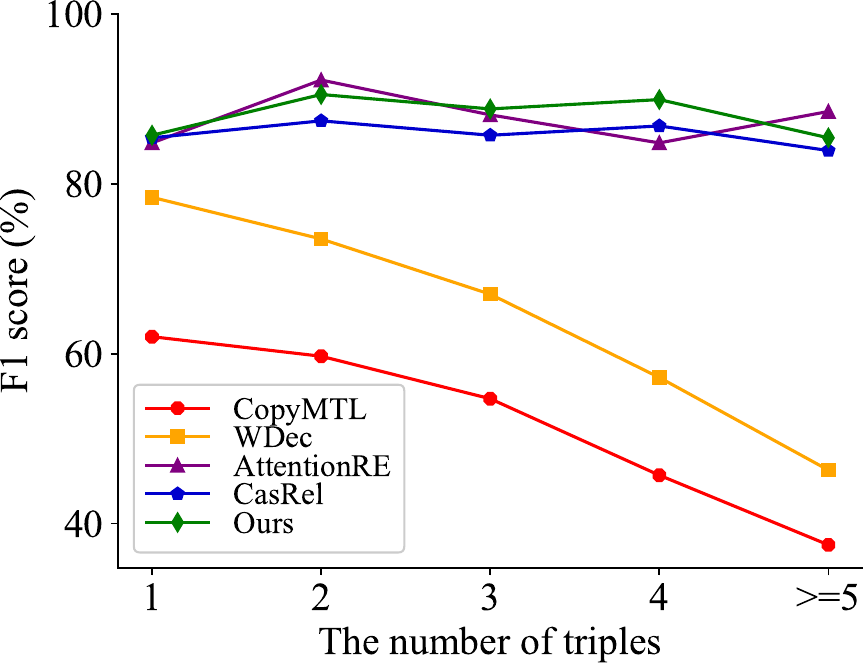}
        \end{minipage}%
        }%
        \subfigure[MDD]{
            \begin{minipage}[t]{0.3\linewidth}
            
           \centering
           \includegraphics[scale=0.3]{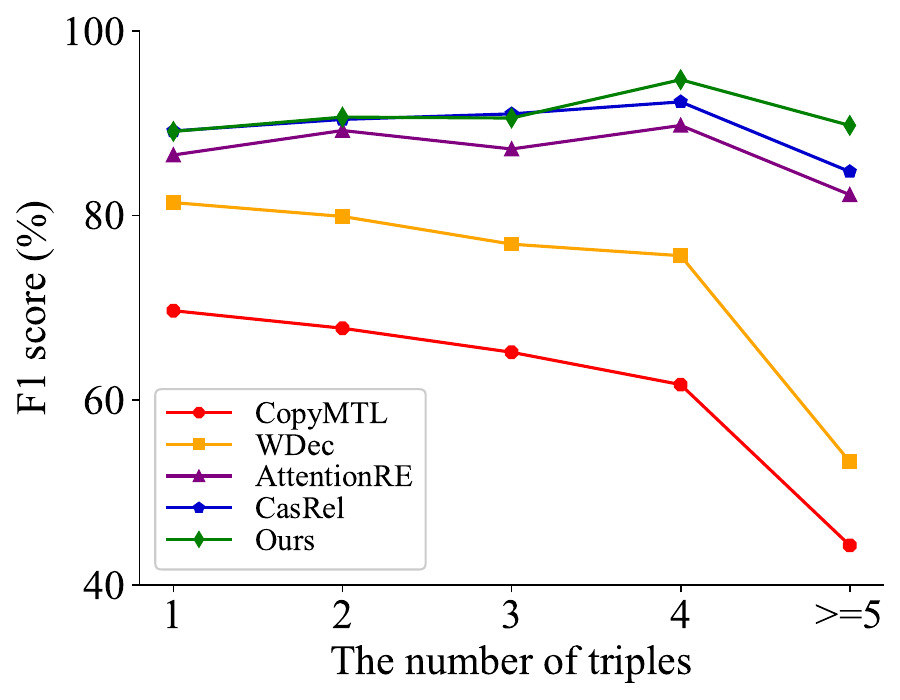}
           \end{minipage}%
           } 
    \caption{ Results of triple extraction from sentences with different number of triples in terms of {\em Exact Match}.}
    \label{fig:nums}
\end{figure*}

Moreover, we conduct an extended test to extract triples from sentences with different number of triples, where the sentence is divided into 5 subclasses, and each class contains texts with 1, 2, 3, 4, or ${\geq}$ 5 triples. Fig. \ref{fig:nums} plots the comparison results of the five methods via the numbers of triples. From the figure, it is shown that our model attains the best F1 scores in most cases on the three datasets, and it performs very stably with the increasing number of triples. Notably, compared with the best baseline CasRel, our model obtains better F1 score on the most difficult class (${\geq}$ 5) on NYT, which indicates that our model is more robust to deal with the complicated scenarios that involve a large number of triples.

\subsubsection{Analysis of Inference Threshold} We report the effect of different thresholds in the text-specific relation decoder in Table \ref{tab:threshold}. As shown, we can obtain the following observation: on NYT and WebNLG, with the increase of the threshold, the recall of our model gradually decreases and the corresponding precision gradually increases. However, the change range of both the precision and the recall is very small.
Specifically, for NYT, as the threshold value increases, the F1 score shows an overall upward trend. When the threshold value is above 0.3, the F1 score is relatively low; otherwise, it is relatively high.
For WebNLG, the F1 score has a downward trend in general with the increase of the threshold value. When the threshold value is below 0.6, the F1 score is relatively high; otherwise, it is relatively low.
One potential reason for the difference in the effect of thresholds on NYT and WebNLG is that WebNLG contains more relations than NYT (see Table \ref{tab:dataset}).

\begin{table*}[t]
\caption{Results on different inference thresholds in terms of {\em Partial Match}.}
\label{tab:threshold}
	\small
\centering
\begin{tabular}{lcccccccccc}
\toprule
\multirow{2}*{Datasets}  & \multirow{2}*{Evaluation indicator}  & \multicolumn{9}{c}{Threshold}\\
\cmidrule(lr){3-11}
                &   &  0.1 & 0.2 & 0.3 & 0.4 & 0.5 & 0.6 & 0.7 & 0.8 & 0.9  \\
\midrule 
  
\multirow{3}*{NYT}   &  precision  & 89.62 & 89.88 & 90.03 & 90.12 & 90.22 & 90.32 & 90.50 & 90.57 & 90.68\\
                        &  recall  & 91.17 &  91.06 & 90.98 & 90.93 & 90.90 & 90.77 & 90.72 & 90.66 & 90.43  \\   
                        &  F1 score  & 90.39 &  90.47 & 90.51 & 90.53 & 90.56 & 90.55 & 90.62  & 90.62 & 90.56\\
\midrule

\multirow{3}*{WebNLG}     &  precision & 89.72 & 89.87 & 90.07 & 90.16 & 90.32 & 90.41 & 90.45 & 90.53 & 90.92    \\ 
                        &  recall   & 92.28 & 92.03 & 91.84 & 91.65 & 91.52 & 91.33 & 91.08 & 90.70 & 89.94\\   
                        &  F1 score  & 90.99 & 90.94 & 90.95 & 90.90 & 90.92 & 90.87 & 90.77 & 90.62 & 90.43     \\
\bottomrule
\end{tabular}
\end{table*}

To sum up, as the threshold changes, the overall performance of our model does not show an obvious change, which also indicates that the threshold has little impact on the performance of our model. The above results again show the robustness of our used TR decoder, and also support the rationality of our design motivation (i.e., first detecting text-specific relation information is effective to guide and enhance the entity extraction process).

\subsubsection{Further Discussion} From the above experimental results, it can be seen that our model obtains the best results in most test cases. However, we also notice that it performs a little worse than its counterpart on WebNLG under {\em Partial Match} (e.g., in Table \ref{tab:res}). This may be because in WebNLG only the last words of entity names are annotated, and sentences are relatively few but with many relations, which may influence our performance. In fact, our method is good at recognizing entity pairs under {\em Exact Match}.
In this section, we give a further analysis of the performance of our model on WebNLG under {\em Partial Match} using a comparison experiment in terms of different sentence types as shown in Table \ref{tab:error_type}. 

Table \ref{tab:error_type} shows the experimental results obtained by our model and CasRel on WebNLG in terms of the two difficult scenarios, which are overlapping patterns ({\em EPO} and {\em SEO}) and different number ($N$) of triples ($N=4$ and $N>=5$). From the table, we can observe that our model shows a stronger ability to deal with extracting overlapping triples from sentences. For example, our model obtains the best F1 score under {\em SEO} ($94.8\%$) , $N=4$ ($92.6\%$) and $N>=5$ ($91.0\%$), respectively. These results further demonstrate that even on WebNLG under {\em Partial Match}, our relation-first cascade approach still shows its merits of being good at dealing with complex overlapping extraction tasks.

\begin{table*}[t]
\caption{Results on difficult scenarios on WebNLG in terms of {\em Partial Match}. Results marked by $\dagger$ are reported from  \cite{DBLP:conf/acl/WeiSWTC20}. }
\label{tab:error_type}
	\small
\centering

\begin{tabular}{lccccc}
\toprule
\multirow{3}*{Methods}  & \multirow{3}*{Evaluation indicator}  &\multicolumn{4}{c}{\em Difficult scenarios}\\
\cmidrule(lr){3-6}
  &  &\multicolumn{2}{c}{Overlapping degree}  &\multicolumn{2}{c}{The number of triples}  \\
\cmidrule(lr){3-4} \cmidrule(lr){5-6}  
  & & SEO & EPO & $=4$ & $>=5$ \\

\midrule

\multirow{1}*{CasRel}   &  F1 score    & 94.7$\dagger$ & 92.2$\dagger$ & 92.4$\dagger$ & 90.9$\dagger$ \\
\midrule

\multirow{1}*{Ours}     &  F1 score   & \bf{94.8} & 92.0 & \bf{92.6} & \bf{91.0}\\
\bottomrule
\end{tabular}

\end{table*}

\section{Conclusion}

In this paper, we propose a joint extraction scheme based on the dual-decoder (i.e., TR decoder and RE decoder) model. The main idea is to first detect relations from text and model them as extra features to guide the entity pair extraction. Our approach is straightforward: the TR decoder detects relations from a sentence at the text level, and then the RE decoder extracts the corresponding head and tail entities for each detected relation. This way, is able to well solve the overlapping triple problem. Experiments on both public datasets and real open pit mine datasets  demonstrate that our proposed model and achieve better F1 scores under the strict evaluation metrics, especially achieving significantly better results on relational triple elements.

The design of our cascade dual-decoder model still has much potential to be improved. Especially, how to enhance the accuracy of extracting candidate relations from the text is a key issue to our relation-first extraction. Note that, the accuracy issue is an open problem for both entity-first and relation-first approaches. That is, for entity-relation, we still have to consider how to ensure accuracy of extracting entities without relation information. This seems like an egg-and-chicken problem. Compared with entity-first approach, our relation-first can make a better use of such information to improve the accuracy of joint extraction. In addition, in the future work, we will pay attention to applying our model in other complex information extraction tasks for open pit mine accident prediction, such as document-level tuple extraction.

\section*{Acknowledgement}
This work was supported in part by the National Natural Science Foundation of China (No. 62333010, 62103150), the National Key Research and Development Program of China (No. 2023YFC2907600), the Key Science and Technology Innovation Project of CCTEG (No. 2021-TD-ZD002, 2022-2-TD-ZD001), the Key Technologies R\&D Program of Liaoning Province (2023JH1/10400082, 2023020456-JH/104), Guangdong Basic and Applied Basic Research Foundation (2024A1515012016).

\appendices


\section{KL Divergence Analysis}\label{app:A}
Our main idea is to minimize the difference between the probability distribution of real triples and the extracted ones from our model, i.e., to minimize our loss function based on  Kullback-Leibler divergence:
\begin{equation}
\min_{\theta}{\underset{x}{\mathbb{E}}}\Big\{KL\big(p((h,r,t)|x) \parallel p_{\theta}((h,r,t)|x)\big)\Big\}.
\end{equation}

We can further deduce the above formula to get the final optimization objective formula as follows:
\begin{align}
  \nonumber &{\underset{\theta}{min}} {\underset{x}{\mathbb{E}}}\Big\{KL\big(p((h,r,t)|x)||p_\theta((h,r,t)|x)\big)\Big\}\\
 \nonumber  =&{\underset{\theta}{min}} {\underset{x}{\mathbb{E}}}
  \Big\{\int_{(h,r,t)}p((h,r,t)|x) \cdot \\& \qquad \qquad \log 
  \frac{p((h,r,t)|x)}{p_\theta((h,r,t)|x)}d(h,r,t)\Big\},\\
  \nonumber \Leftrightarrow&{\underset{\theta}{min}} {\underset{x}{\mathbb{E}}}
  \nonumber \Big\{\int_{(h,r,t)}-p((h,r,t)|x)\cdot \\
  \nonumber &\qquad  \qquad  \log{p_\theta((h,r,t)|x)}d(h,r,t)\Big\}\\
  \nonumber =&{\underset{\theta}{min}} \int_{x}p(x)\Big[\int_{(h,r,t)}-p((h,r,t)|x)\cdot \\
  \nonumber &\qquad \qquad \qquad \log p_{\theta}((h,r,t)|x)d(h,r,t)\Big]dx\\
  \nonumber =&{\underset{\theta}{min}}-{\underset{(h,r,t),x}{\mathbb{E}}} \Big\{\log p_{\theta}((h,r,t)|x) \Big\}\\
  =&\nonumber \min_{\theta} - {\underset{(h,r,t),x}{\mathbb{E}}}\Big\{\log \big{(}p_{\theta}(r|x)\cdot p_{\theta}(h|r,x) \cdot\\
&\qquad \qquad \qquad \quad p_{\theta}(t|r,h,x)\big{)}\Big\} \label{eq:min} .
\end{align}

Eq. \ref{eq:min} straightforwardly guides the design of our model structure.

\section{Posterior Probability Analysis}\label{app:B}
Based on Eq. \ref{eq:min}, our model structure has two parts: one is the text-specific relation decoder corresponding to $p_{\theta}(r|x)$, and the other is relation-corresponded entity decoder corresponding to $p_{\theta}(h|r,x)$ and $p_{\theta}(t|r,h,x)$. However, both probabilities in the relation-corresponded entity decoder are posterior ones conditioned on the text-specific relation $r$. It is necessary to explore whether extra information $r$ is more effective to improve the accuracy, i.e., the posterior probability is better.

To illustrate the effectiveness of adding the text-specific relation $r$ as extra information to predict head entities and tail entities, we give the following theoretical analysis:


Since the extraction process of tail entities is similar to that of head  ones, we only need to investigate the case of head entity extraction. The function between relations $r$ and head entities $h_k$ is defined as 
\begin{align}
r=f(h_k)+e,
\end{align}
where $e$ denotes {\em Gaussian white noise} , $f$ is a complex nonlinear function. We assume that $r$, $h_k$, and $e$ respectively obey the following Gaussian distribution:
\begin{align}
&h_k \sim \mathcal {N}(m_h, P_{11}),\\
&e \sim \mathcal {N}(0, {\sigma }^2),\\
&r \sim \mathcal {N}(m_r, P_{22})\qquad(P_{22}>0)\label{r}.
\end{align}
Therefore, without extra information $r$, the best estimate of head entities is $m_h = \mathbb{E}\{h_k\} = \hat{h}_k$, which is the one as used in \cite{DBLP:conf/acl/WeiSWTC20}.

Turning now to the joint distribution of $h_k$ and  $r$, we assume that $h_k$ and $r$ are jointly normally distributed, and the distribution is given as follows:
\begin{align}
\binom{h_k}{r} \sim \mathcal {N}\begin{pmatrix}
{\begin{bmatrix}
m_h\\ 
m_r
\end{bmatrix}}
,&
{\begin{bmatrix}
P_{11} & P_{12}\\ 
P_{21} & P_{22}
\end{bmatrix}}
\end{pmatrix}  \label{hr}.
\end{align}

A new random variable $h_{k+1}$ is defined as the new measurement of head entities $h_k$ given the condition of extra information $r$. According to the Bayes formula, we have the posterior probability as follows:
\begin{align}
    p(h_{k+1}) = p(h_k|r) = \frac{p\binom{h_k}{r}}{p(r)}.
\end{align}

Due to the marginal distribution and properties of normal distribution, the random variable $h_{k+1}$ is also {\em Gaussian}, and can be easily calculated as follows:
\begin{align}
    h_{k+1} \sim \mathcal {N}(m_{h|r}, P),
\end{align}
where $m_{h|r} = \mathbb{E}\{h_{k|r}\} = \mathbb{E}\{h_{k+1}\} = \hat{h}_{k+1}$.
Thus, the least squares estimate of head entities, given relations, is calculated as
\begin{align}
\nonumber &\hat{h}_{k+1}=m_h+P_{12}P_{22}^{-1}(r-m_r)\\ 
&\qquad\ =\hat{h}_{k}+P_{12}P_{22}^{-1}(r-m_r).  \label{hk1} 
\end{align}

Note that, the error $\tilde{h} = h_{k+1}-\hat{h}_{k+1}$ provides significant information about the accuracy of our algorithm. Indeed the mean square estimation error and the expectation error are respectively given by the following formulae:
\begin{align}
 \nonumber&trace(\mathbb{E}\{\tilde{h}*\tilde{h}^T\}) = trace(Cov\{h_{k+1}\}) \\
&\qquad \quad \quad  \quad  \qquad \ \ = trace(P), \\
&\mathbb{E}\{{\tilde{h}}\} = \mathbb{E}\{h_{k+1}\}-\mathbb{E}\{\hat{h}_{k+1}\} = 0.
\end{align}

Therefore, it is obvious that our algorithm is superior to the one in \cite{DBLP:conf/acl/WeiSWTC20} by the following formulae, that is, extra information $r$ is beneficial to the prediction of head entities (and also tail entities). Experiments on relational triple elements also validate our theoretical analysis.
\begin{align}
     \nonumber \mathbb{E}\{{\tilde{h}}\}=&\mathbb{E}\{{(h_{k+1}-\hat{h}_k)-(\hat{h}_{k+1}-\hat{h}_k)}\}\\
    =&\mathbb{E}\{{h_{k+1}-\hat{h}_k}\}-\mathbb{E}\{{\hat{h}_{k+1}-\hat{h}_k}\}=0, \\
\Rightarrow&
    \nonumber \mathbb{E}\{{h_{k+1}-\hat{h}_k}\}= \mathbb{E}\{{\hat{h}_{k+1}-\hat{h}_k}\} \\
    \nonumber =& \hat{h}_{k+1}-\hat{h}_{k} \\
     = &P_{12}P_{22}^{-1}(r-m_r)\neq0.
\end{align}

Note that we do not consider the condition of $x$, since both sides have the same condition, which is equivalent to without this condition. Alternatively, we can add this condition and rederive the formulae, which will lead to the same conclusion.

\begin{table*}[t]
\caption{Results on the parallel relation and cascade relation between head-entity extractor and tail-entity extractor.}
\label{tab:parallel}
	\small
\centering

\begin{tabular}{lccc}
\toprule
    \multirow{2}[0]{*}{Methods} & NYT   & WebNLG & MDD \\
\cmidrule(lr){2-4}
          & F1    & F1    & F1 \\
\midrule
    Parallel realtion & 87.6  & 85.7  & 69.19 \\
    Cascade realtion (ours) & 90.6  & 88.4  & 72.36 \\
\bottomrule
    \end{tabular}%

\end{table*}

In addition, in order to verify the effectiveness of the cascade relationship between the head entity extractor and the tail entity extractor, we also conducted experiments with the parallel relationship between the head entity extractor and the tail entity extractor. As shown in Table \ref{tab:parallel}, the experimental results proved that the cascade relation is more effective than the parallel relation in extracting overlapping triple relations. For example, on the NYT dataset under {\em Exact Match}, the F1 score is 87.6\% obtained using the parallel relation approach, and the F1 score decreases by 3.0\% compared to the cascade relation approach. On the WebNLG dataset under {\em Exact Match}, the F1 score is 85.7\% obtained using the parallel relation approach, and the F1 score decreases by 2.7\% compared to the cascade relation approach. On the MDD dataset, the F1 score is 69.19\% obtained using the parallel relation approach, and the F1 score decreases by 3.17\% compared to the cascade relation approach.



\ifCLASSOPTIONcaptionsoff
  \newpage
\fi

\bibliographystyle{IEEEtran}

\bibliography{Manuscript}

\begin{IEEEbiography}[{\includegraphics[width=1in,height=1.25in,clip,keepaspectratio]{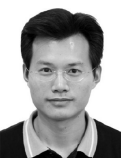}}]{Jian Cheng} received the B.Sc. degree in Automation, the M.Sc. degree in Control Theory and Control Engineering, and the Ph.D. degree in Communication and Information System from the China University of Mining and Technology, Xuzhou, China, in 1997, 2003, and 2008 respectively. He has been a postdoctoral fellow at Tsinghua University and University of Birmingham from 2009 to 2013. He is currently a Professor and the Chief Scientist with the Research Institute of Mine Artificial Intelligence in Chinese Institute of Coal Science (Tiandi Science and Technology Co., Ltd.), State Key Laboratory for Intelligent Coal Mining and Strata Control, Beijing, China. His current research interests include machine learning and pattern recognition, data mining and big data, as well as imbalance learning and image processing and their applications in industrial fields.\end{IEEEbiography}

\vspace{-10 mm}

\begin{IEEEbiography}[{\includegraphics[width=1in,height=1.25in,clip,keepaspectratio]{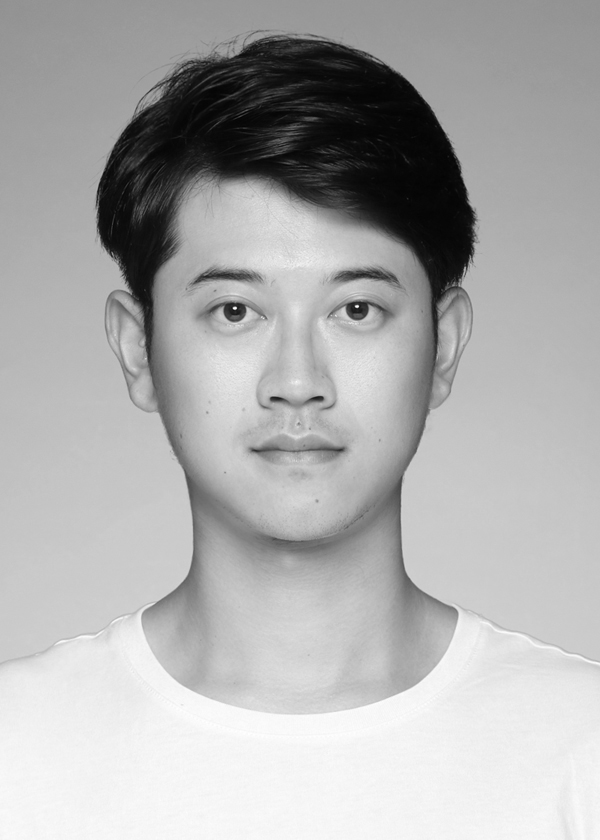}}]{Tian Zhang} received the B.Sc. degree in Digital Media Technology from Shenyang Institute of Engineering, Shenyang, China, in 2018, and the M.Sc. degree from Northeastern University, Shenyang, China,in 2021. He is currently pursuing the Ph.D degree with the College of Software, Northeastern University, Shenyang China. His research interests include computational intelligence and knowledge graph.\end{IEEEbiography}

\vspace{-10 mm}

\begin{IEEEbiography}[{\includegraphics[width=1in,height=1.25in,clip,keepaspectratio]{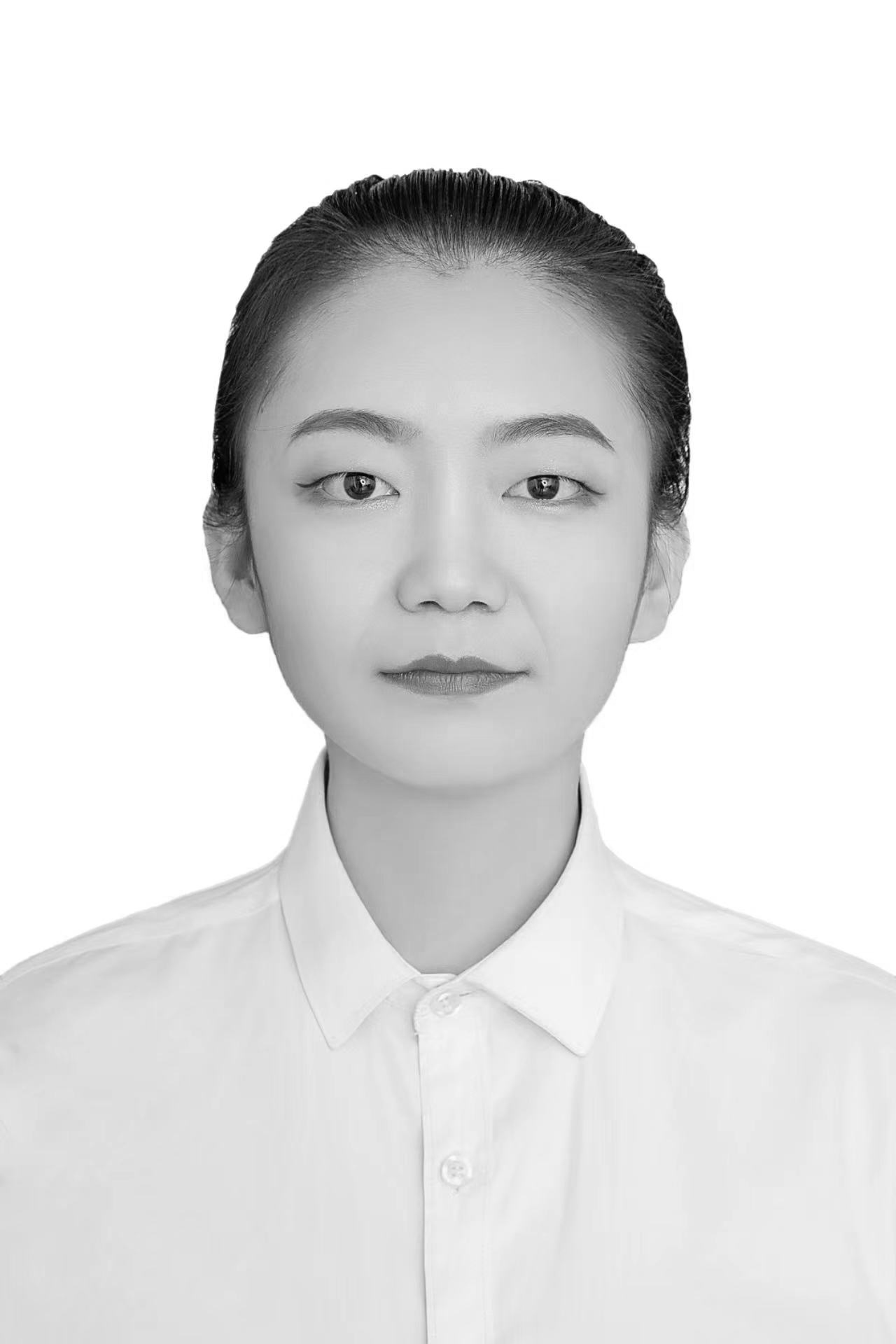}}]{Shuang Zhang} received her master's degree from Northeastern University in China in 2023. Her main research area is knowledge graphs, with a focus on named entity recognition and relationship extraction tasks.\end{IEEEbiography}

\vspace{-10 mm}

\begin{IEEEbiography}[{\includegraphics[width=1in,height=1.25in,clip,keepaspectratio]{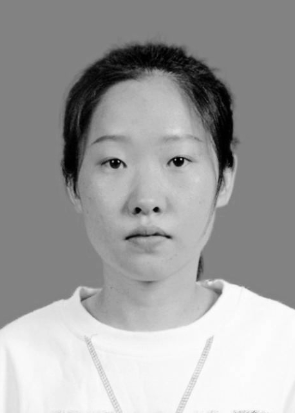}}]{Huimin Ren} received the B.Sc. degree in software engineering from Shanxi University, Taiyuan, China, in 2019, and received her master's degree from Northeastern University, Shenyang, China in 2022. Her main research area is knowledge graphs, with a focus on named entity recognition and relationship extraction tasks.\end{IEEEbiography}

\vspace{-5 mm}

\begin{IEEEbiography}[{\includegraphics[width=1in,height=1.25in,clip,keepaspectratio]{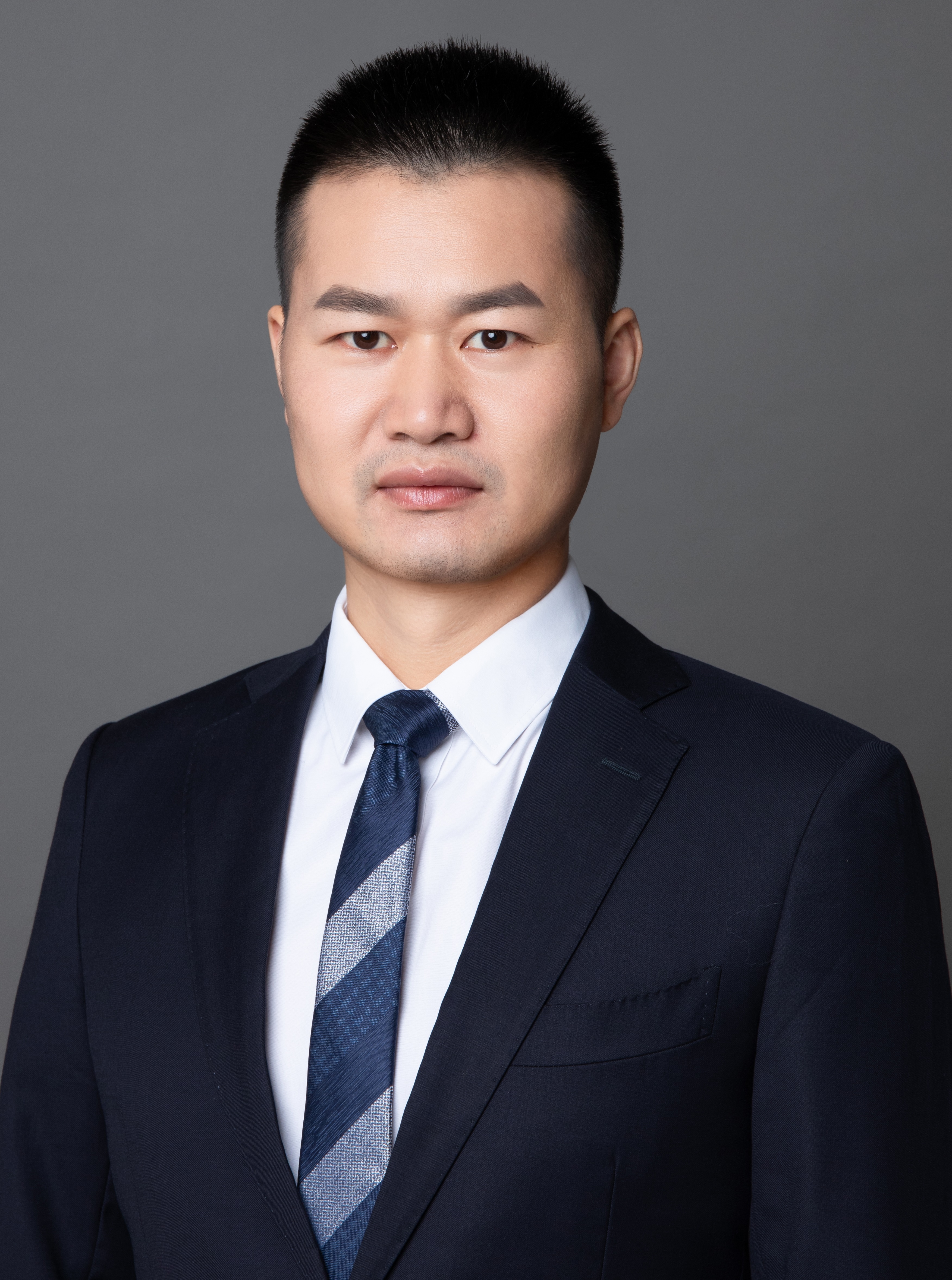}}]{Guo Yu}(Member, IEEE) received the B.S. degree in information and computing science and the M.Eng. degree in computer technology from Xiangtan University, Xiangtan, China, in 2012 and 2015, respectively. He received the Ph.D. degree in computer science from University of Surrey, Guildford, U.K., in 2020.   From 2020 to 2022,  he was a research fellow with the Key Laboratory of Smart Manufacturing in Energy Chemical Process, East China University of Science and Technology, Shanghai, China. 
    
He is currently an associate professor in Nanjing Tech University. He servers as the editor board member of  Evolutionary Computation. His current research interests include artificial intelligence and machine learning. He is a regular reviewer of more than 10 journals and conferences such as the IEEE TEVC, IEEE TCYB, TNNLS, TFS, TETCI, CAIS, and CVPR.\end{IEEEbiography}

\vspace{-5 mm}

\begin{IEEEbiography}[{\includegraphics[width=1in,height=1.25in,clip,keepaspectratio]{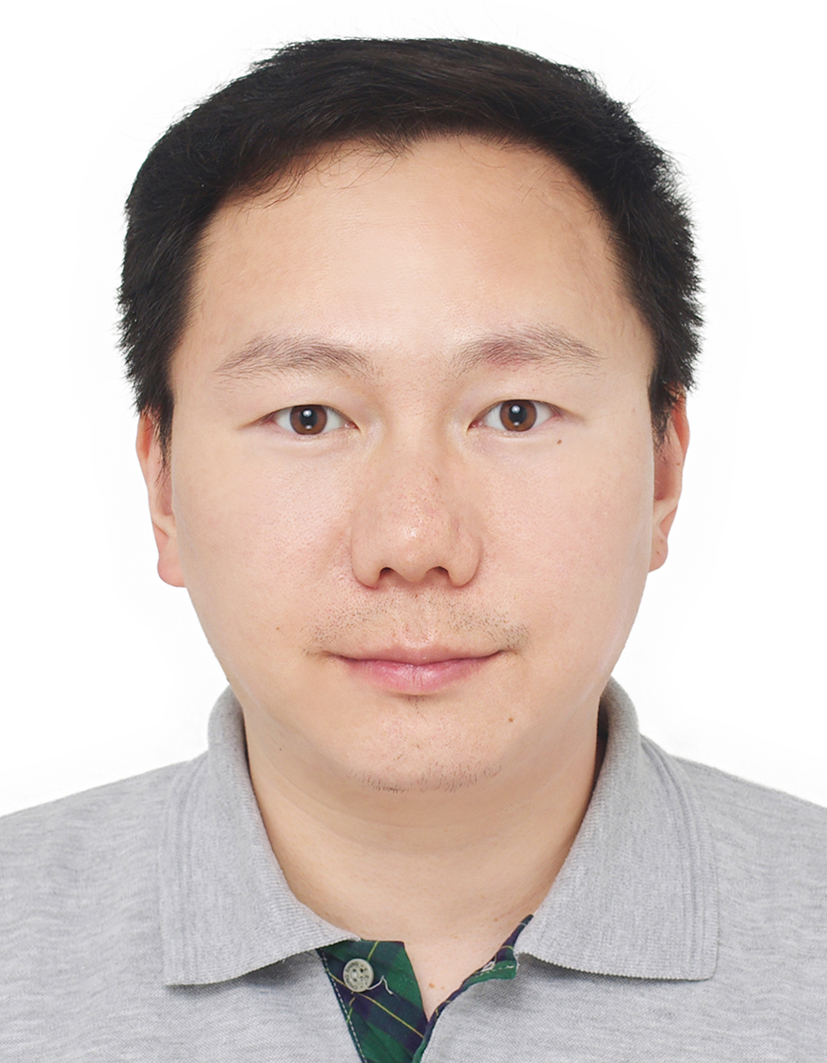}}]{Xiliang Zhang}(Member, IEEE) received the B.E. degree in Automatic Control from Northeastern University, Shenyang, China, in 2007, the M.S. degree in Control Systems from Imperial College London, London, UK, in 2009, and the Ph.D. degree in Automatic Control and Systems Engineering from the University of Sheffield, UK, in 2016. Since 2017, he has been a lecturer with School of Intelligent Manufacturing and Control Engineering, Shanghai Polytechnic University.\end{IEEEbiography}

\vspace{-5 mm}

\begin{IEEEbiography}[{\includegraphics[width=1in,height=1.25in,clip,keepaspectratio]{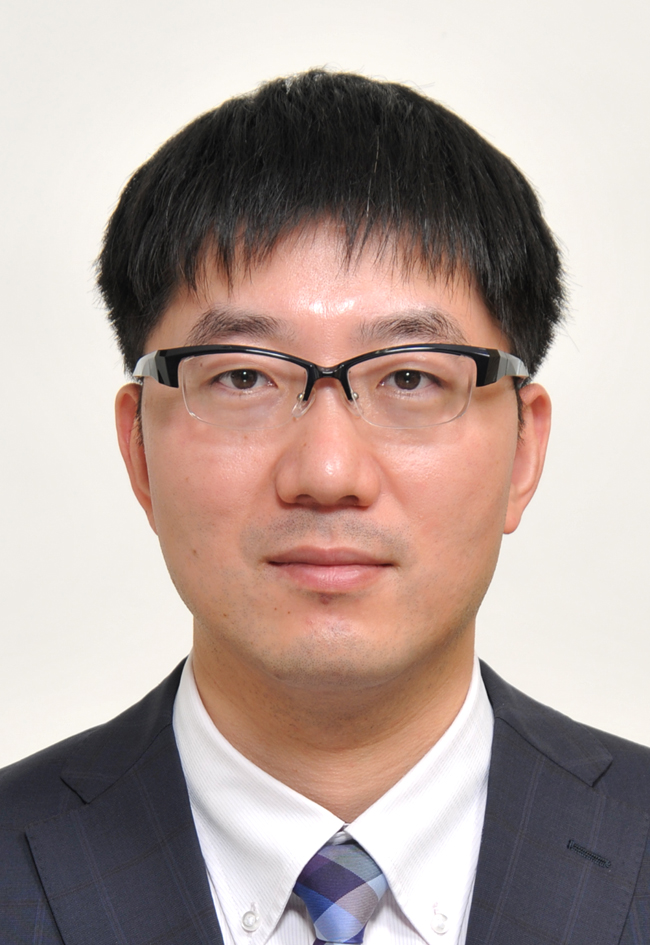}}]{Shangce Gao}(Senior Member, IEEE) received the Ph.D. degree in innovative life science from the University of Toyama, Toyama, Japan, in 2011. He is currently a Professor with the Faculty of Engineering, University of Toyama. His current research interests include nature-inspired technologies, machine learning, and neural networks for real-world applications. His research has led to over 150 publications in top venues. He serves as an Associate Editor for many international journals, such as IEEE Transactions on Neural Networks and Learning Systems and IEEE/CAA Journal of Automatica Sinica. He has served on the program committees for several international professional conferences, including AAAI, NeurIPS, CVPR, and IEEE CEC.\end{IEEEbiography}

\vspace{-5 mm}

\begin{IEEEbiography}[{\includegraphics[width=1in,height=1.25in,clip,keepaspectratio]{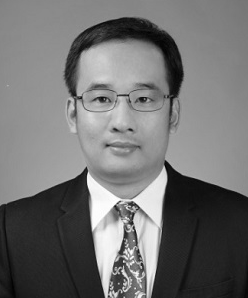}}]{Lianbo Ma}(Senior Member, IEEE) received the B.Sc. degree in communication engineering and the M.Sc. degree in communication and information systems from Northeastern University, Shenyang, China, in 2004 and 2007, respectively, and the Ph.D. degree from the University of Chinese Academy of Sciences, Beijing, China, in 2015.

He is currently a Professor with Northeastern University. He has published over 90 journal articles, books, and refereed conference papers. His current research interests include computational intelligence and machine learning.\end{IEEEbiography}

\vfill

\end{document}